\documentclass{article}

\usepackage{microtype}
\usepackage{graphicx}
\usepackage{booktabs} 

\usepackage{hyperref}

\usepackage[accepted]{icml2025}

\usepackage{amsmath}
\usepackage{amssymb}
\usepackage{mathtools}
\usepackage{amsthm}

\usepackage[capitalize,noabbrev]{cleveref}

\theoremstyle{plain}

\theoremstyle{definition}

\theoremstyle{remark}

\usepackage[textsize=tiny]{todonotes}

\usepackage{amsmath,amssymb,amsfonts,amsthm}
\usepackage{graphicx}
\usepackage{textcomp}
\usepackage{xcolor}
\usepackage{booktabs}
\usepackage{multirow}
\usepackage{multicol}
\usepackage{bm}
\usepackage{comment}
\DeclareMathOperator*{\argmax}{\arg\!\max} 
\usepackage{atbegshi}
\usepackage{subcaption}

\usepackage{lipsum}
\usepackage{graphicx}
\usepackage{caption}

\icmltitlerunning{Percentile-Based Deep Reinforcement Learning and Reward Based Personalization For Delay Aware RAN Slicing in O-RAN}

\begin{document}

\twocolumn[
\icmltitle{Percentile-Based Deep Reinforcement Learning and Reward Based Personalization For Delay Aware RAN Slicing in O-RAN}




\begin{icmlauthorlist}
\icmlauthor{Peyman Tehrani}{yyy}
\icmlauthor{Anas Alsoliman}{yyy}
\end{icmlauthorlist}

\icmlaffiliation{yyy}{Donald Bren School of Information and Computer Sciences, University of California at Irvine, United States}

\icmlcorrespondingauthor{Peyman Tehrani}
{peymant@uci.edu}

\icmlkeywords{Deep reinforcement learning, Federated learning, Personalization, RAN slicing, Wireless networks, Reward function, Resource allocation, Probabilistic constraint, Percentile, Quality of Service.}

\vskip 0.3in
]



\printAffiliationsAndNotice{} 

\begin{abstract}

In this paper, we tackle the challenge of radio access network (RAN) slicing within an open RAN (O-RAN) architecture. Our focus centers on a network that includes multiple mobile virtual network operators (MVNOs) competing for physical resource blocks (PRBs) with the goal of meeting probabilistic delay upper bound constraints for their clients while minimizing PRB utilization. Initially, we derive a reward function based on the law of large numbers (LLN), then implement practical modifications to adapt it for real-world experimental scenarios. We then propose our solution, the Percentile-based Delay-Aware Deep Reinforcement Learning (PDA-DRL), which demonstrates its superiority over several baselines, including DRL models optimized for average delay constraints, by achieving a 38\% reduction in resultant average delay. Furthermore, we delve into the issue of model weight sharing among multiple MVNOs to develop a robust personalized model. We introduce a reward-based personalization method where each agent prioritizes other agents' model weights based on their performance. This technique surpasses traditional aggregation methods, such as federated averaging, and strategies reliant on traffic patterns and model weight distance similarities.

\end{abstract}

\section{Introduction} 

\subsection{Motivation}

The field of wireless communication has seen tremendous advancements in recent years, leading to the emergence of 5G networks that promise to offer unprecedented speeds, reliability, and capacity. However, with the increasing number of devices and applications that require connectivity, the demand for more efficient and flexible radio access network (RAN) management has become a pressing issue. Thus, mobile network operators are expected to support diverse range of applications where each of them has its own specifc requirements and constraints. These applications range from mission-critical applications like autonomous driving and remote surgery, necessitating high reliability and low latency communication links, to high-bandwidth demanding applications such as virtual reality and 4K video streaming. Furthermore, the rise of internet of things (IoT) and smart city initiatives demands networks that can handle long-range communications with numerous active devices. In response to the escalating demands of evolving technologies, mobile network operators (MNO) are veering toward solutions like network and RAN slicing \cite{wijethilaka2021survey,elayoubi20195g}. These methods enable the partitioning of a single network into multiple tailored network slices, ensuring each segment meets the specific requirements of its use-case. This tailored approach not only reduces operational costs but significantly improves the overall quality of service (QoS).

Network and RAN slicing is a well studied problem in recent years, as many work considered optimization and model based approachs \cite{motalleb2022resource,popovski20185g}. However, the industry's shift towards open radio access networks (O-RAN) \cite{polese2023understanding,bonati2021intelligence} can provide new possibilities to the field. Integrating artificial intelligence (AI) and machine learning (ML), particularly in the RAN intelligent controller (RIC) component, paves the way for a data-driven, closed-loop control system. Utilizing the vast amounts of data generated in the network every second, these data-driven approaches can significantly boost the performance and energy efficiency of the future generation of network infrastructure. Unlike the static model based methods, the learning-based approaches offer dynamic adaptability to network environments, enabling MNOs to develop specialized ML models. These models can be tailored to specific regions and use cases, taking into account unique factors such as user traffic patterns, level of QoS, wireless propagation of geographical areas, and other relevant environmental variables. 
\subsection{Main Contributions}
In this paper we study the RAN slicing problem in an O-RAN environment. Specifically, we consider a physical resource block (PRB) allocation problem for multiple mobile virtual network operators (MVNOs) in a heterogeneous setting, where the client traffic demands, number of users, required QoS, and wireless propagation environment for each MVNO could be completely different among the other MVNOs. The complexity and diversity of these environmental variables necessitate a specialized model capable of dynamically adapting to each MVNO's setting. To address this, we utilize a data-driven approach, specifically employing deep reinforcement learning (DRL) algorithms. These algorithms have shown notable efficacy in managing complex control tasks in challenging settings \cite{mnih2015human}. Additionally, we propose a reward function based on the law of large numbers (LLN) to satisfy the probabilistic upper bound delay constraint of the MVNOs's clients data request.

We propose a Percentile-based Delay Aware DRL (PDA-DRL) solution and demonstrate its superiority over a diverse set of baselines. We further explore how multiple MVNOs can collaboratively share their model weights without disclosing any user data to improve their policy performance. We demonstrate that federated learning baselines such as federated averaging (FedAV) \cite{mcmahan2017communication} will not work in this setting as there is a huge difference between the environments that each agent interacts with. Instead, we propose a reward-based personalization method in which each agent up-weights the other agents' models based on their estimated performance on their own environment. We also validate our approach through online DRL experiments using our experimental framework, which incorporates the Colosseum, the world's largest RF emulator \cite{bonati2021colosseum}, and Scope, a next-generation wireless network prototyping platform \cite{bonati2021scope}.

In summary, our contributions are multifaceted and distinct from other works reviewed in Section \ref{sec:litreview}, and can be summarized as follows:

\begin{itemize}
\item \textbf{Percentile-Based Delay Guarantees}: Our work explores percentile-based delay guarantees as a more robust metric than the average delay or throughput, which are commonly discussed in the literature. 

\item \textbf{Reward Shaping Function}: We introduce a distinctive reward shaping function, built on top of the RAN slicing constrained optimization solution. 

\item \textbf{Performance Based Personalization Method}: Our personalization approach is novel, focusing on the performance of agents rather than similarities like traffic patterns or model weights. With the increasing accessibility of digital twins \cite{ericsson2023randigitaltwins,O-RAN2024} and generative AI \cite{karapantelakis2023generative}, our proposed method would be practical and implementable by network operators.

\item \textbf{Validation on Experimental Testbed}: Unlike most works in Section \ref{sec:litreview} that rely solely on simulation, we validate our DRL solution using the Colosseum experimental testbed.

\end{itemize}

\section{Literature Review} \label{sec:litreview}

Recently, the deployment of ML in O-RAN has garnered significant attention for a variety of use cases, including energy saving \cite{akman2024energy}, traffic steering \cite{lacava2022programmable}, resource allocation \cite{joda2022deep}, automation \cite{d2022orchestran}, traffic prediction \cite{niknam2022intelligent}, anomaly detection \cite{sun2024spotlight}, access control \cite{cao2021user}, load balancing \cite{orhan2021connection}, remote electrical tilt optimization \cite{vannella2022off}, \cite{orhan2021connection}, admission control \cite{lien2021session}, energy efficiency \cite{kalntis2022energy}, security \cite{xavier2023machine}, spectrum sensing \cite{reus2023senseoran}, virtual reality \cite{kougioumtzidis2023deep}, MU-MIMO interference coordination \cite{ge2023learning} and many others. In the context of RAN slicing, several notable works have been conducted and here we will provide a review of the most related studies.

Bakri et al. \cite{bakri2021data} utilize traditional ML algorithms like support vector machines for predicting radio resource requirements in network slices, based on user channel quality feedback. Filali et al. \cite{filali2022dynamic} propose a two-level RAN slicing method within the O-RAN framework, focusing on the allocation of communication and computation resources using a double deep Q-network (DDQN) algorithm. Yang et al. \cite{yang2021ran} address RAN slicing for IoT and URLLC services, formulating it as an optimization problem and proposing a solution via the alternating direction method of multipliers (ADMM). Hua et al. \cite{hua2019gan} introduce a generative adversarial network-powered deep distributional Q Network (GAN-DDQN) for resource management in network slicing. Setayesh et al. \cite{setayesh2022resource} and Wu et al. \cite{wu2020dynamic} explore multi-timescale RAN slicing problems, proposing hierarchical deep learning frameworks and studying RAN slicing for Internet of Vehicles (IoV) services, respectively. Mei et al. \cite{mei2021intelligent} focus on a two-layered RAN slicing strategy aimed at maximizing QoS and spectrum efficiency.

However, none of the previous works consider a percentile-based delay guarantee or probabilistic QoS constraint. Moreover, the aforementioned studies primarily focus on single wireless environments and do not address collaborative efforts among different agents in varied and diverse environments and tasks. Few works, such as \cite{messaoud2020deep,tehrani2021federated,abouaomar2022federated} has studied aggregating multiple DRL agents. Nonetheless, these approaches mainly employ federated averaging \cite{mcmahan2017communication}, which, as our study demonstrates, is less effective in environments with significant variations in traffic patterns, wireless propagation, and task-specific QoS levels.

Among the limited studies that focus on personalization for each DRL agent's model, Rezazadeh et al. \cite{rezazadeh2022specialization} propose a dynamic user clustering method based on similarities in local traffic conditions. However, their approach does not account for service type and QoS level, reducing its applicability in scenarios with diverse QoS constraints. Nagib et al. \cite{nagib2023accelerating} explore the use of transfer learning to accelerate RL-based RAN slicing convergence, suggesting a predictive method to choose the most appropriate pre-trained policy by focusing on convergence error and weight vector distances. In contrast, our research points out the inadequacies of relying solely on model weight similarities, demonstrating  that personalization based on model performance is more effective.

\section{System Model} \label{sec:sysmdl}

\begin{figure}[t!]
\centering
        \includegraphics[width=.4\columnwidth]{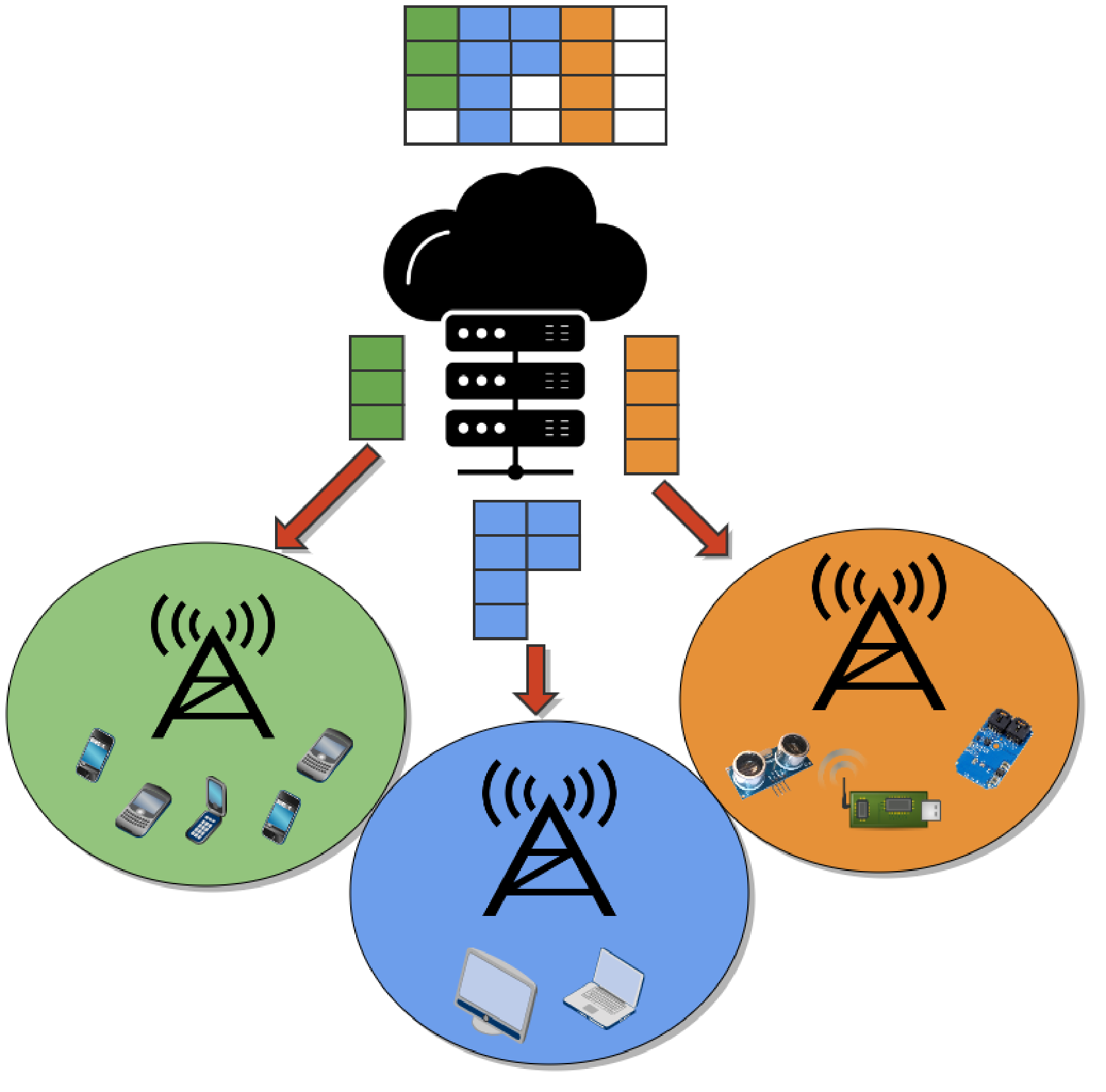}
        \caption{A network containing a central controller (telco operator) and multiple RANs  with their associated users.}
        \label{fig:sysmdl}
\end{figure}

We consider a network composed of a core network (CN) and  multiple RANs, each RAN is represented as a cell tower providing radio access to CN. We assume that both the CN and the RANs are owned and managed by a single telco operator (TO) which provides radio network resources, represented as PRBs, to MVNOs, who periodically rent a slice of RAN resources to provide cellular network access to their own mobile users as illustrated in Fig. \ref{fig:sysmdl}. Therefore in each cell, multiple MVNOs would compete for physical RAN resources based on their clients demand and the QoS that they have guaranteed to their users.  

To be more precise, assume that there exists $K$ different RAN clusters (or we can call it cell) which we index it with $ k \in \mathcal{K}= \{1,2,...,K\}$ and there are total of $V$ MVNOs as we index it by $ v \in \mathcal{V}= \{1,2,...,V\}$ . We show the user $j$ in cell $k$ which is client of MVNO $v$ as $j^v_k$. We also assume each MVNO has its own set of strict QoS requirements. Here, we focus on maximum transmission latency as a QoS metric perceived by the operator's clients. We denote the maximum transmission latency guaranteed by the $v$th MVNO as $D^v_{max}$.

We also assume each user has a specific data request distribution, where the number of requests for user $j$ in cell $k$ is drawn from $Q_{k,j}$ and the data size from $F_{j,k}$. For example, sensory clients generate many small packets, while large file downloads result in fewer, larger packets. These distributions vary based on protocols, user types, and timing, and the request arrivals may not follow predefined models known to the MVNOs.

At each slicing time slot $T_S$, each MVNO determines the number of PRBs required to meet its users' demands, in line with anticipated QoS. The quantity of requested PRBs depends on factors like latency requirements, wireless channel quality, data request queue length, the monetary value of resource blocks, and other environmental variables. Here, PRB refers to a time-frequency network resource block, with a time duration of one PRB being $T_{B}[s]$ time-width and $W[Hz]$ frequency-width. We also  define two distinct timing scales: the PRB time slot $T_{B}$ and the slicing time slot $T_{S}$. The $T_S$ is greater than $T_B$ and we assume it is an integer multiplicative of $T_B$ as $T_{S}=H T_B$. In our slicing framework, at the start of each $T_S$, each MVNO requests a specific amount of PRBs from TO. Upon allocation, a scheduling algorithm distributes the assigned PRBs to mobile users at the $T_B$ time scale in order to meet the required QoS.

\begin{figure}[t!]
\centering
        \includegraphics[width=.75\columnwidth]{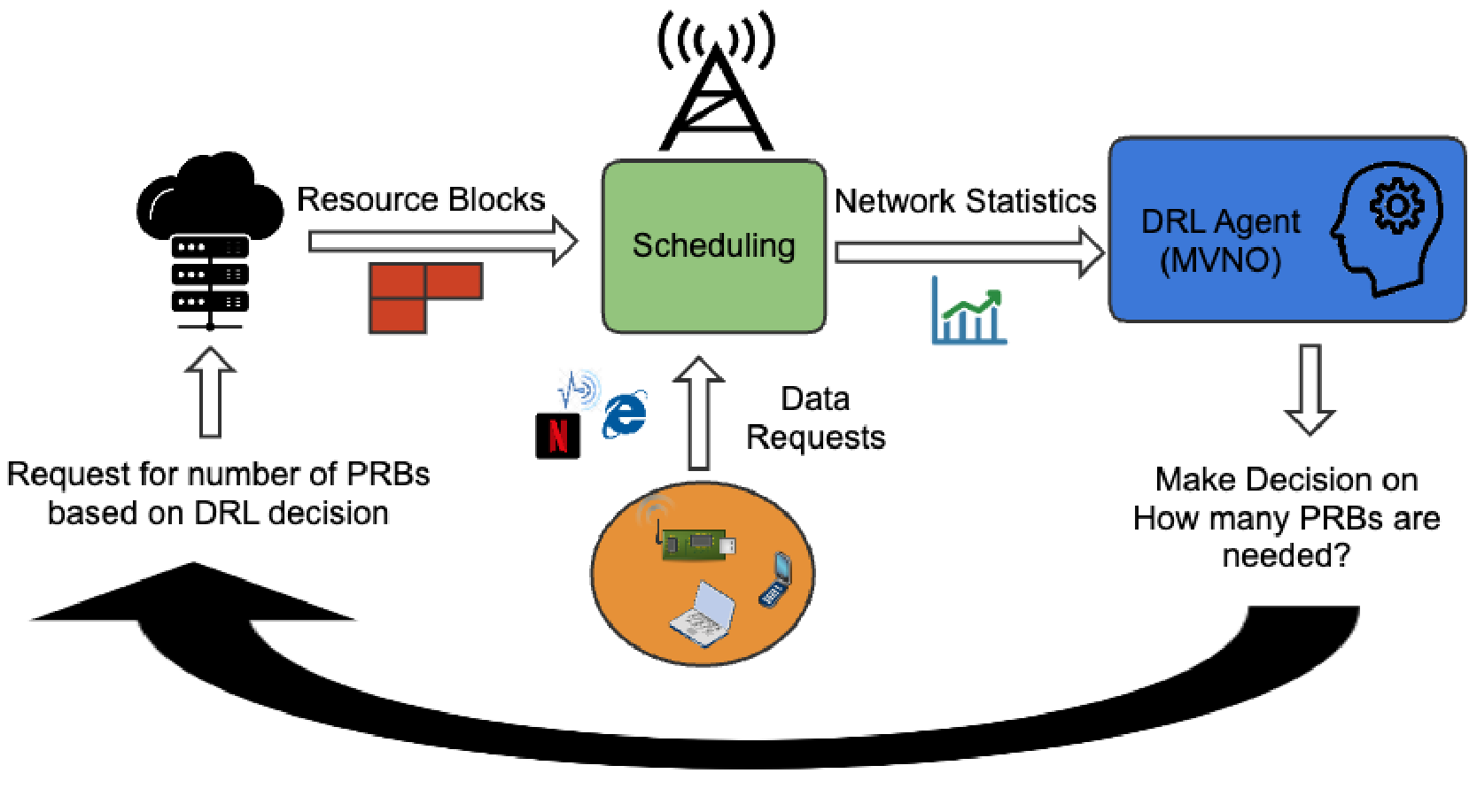}
        \caption{RL scheme for the RAN slicing problem.}
        \label{fig:DQN}
\end{figure}

Due to the nature of the queuing and existence of unknown traffic patterns, and the fact that current actions directly affect future states and actions, this problem is a sequential decision-making problem. One appropriate tool for solving this type of problem is DRL. Essentially, we can model the problem as a two-layer optimization problem. At the higher level, the MVNO acts as an DRL agent that needs to decide at every $T_S$ seconds, how many PRBs to request from the TO in order to satisfy both its near-future traffic demands and its monetary budget. At the lower level, the MVNO must satisfy its users' latency requirements over the next $HT_B$ time slots by solving a scheduling problem using the assigned PRBs. This procedure is summarized in Fig. \ref{fig:DQN}.

\section{Problem Formulation}\label{sec:probform}

The downlink SNR of user $j$ in cell $k$ at time slot $t$ can be defined as:

\begin{equation}
 \gamma^{t}_{j,k}=\frac{P^{t}_k g^t_{j,k}}{N_j}
\end{equation}
where $g^t_{j,k}$ is the channel gain between the kth RAN and jth
user at time slot $t$,  $P_t$ is the $k$th RAN’s RF transmit power and $N_j$ is the noise power at the user $j$ terminal. It is assumed that the each RAN is using uniform power allocation for all PRBs. We also assume that the channel gain $g^t_{j,k}$ includes both small-scale and large-scale fading factors, and it remains constant for a period of length $T_b$.

Each RAN use a predefined mapping lookup table based on the channel quality indicator (CQI) reported by the mobile users, which maps the measured SNR at user $j$ to an integer bit rate capacity which is related to a modulation and coding scheme (MCS) as follows $\Psi : \mathcal{R} \rightarrow \mathcal{N} $. The number of deliverable bits for the $j$th user request in the requests’ buffer for $k$th PRB at time $t$ would be:

\begin{equation}
 C^t_{j,k}=WT_{b} \Psi(\gamma^{t}_{j,k})
\end{equation}

We define $f^{t}_{j^v_k,q}$, as the $q$th packet size requested in cell $k$ by user $j$ for operator $v$ at time $t$. At each PRB time slot, $C^t_{j,k}l^t_{j^v_k,q}$ bits will be delivered to user $j$, where $l^t_{j^v_k,q}$ is the number of PRBs allocated to the $q$th request of  user $j$ by operator $v$. Therefore given that the guaranteed transmission latency for $v$th operator is $D^v_{max}$, then the following constraint should be satisfied:

\begin{equation}
\sum_{t=t_0} ^{t_0+d_{j^v_k,q}} C^t_{j,k}l^t_{j^v_k,q} \geq f^{t_0}_{j^v_k,q}
\label{eq:delvier_const}
\end{equation}
\begin{equation}
d_{j^v_k,q} \leq D^v_{max}
\label{eq:maxdelay_const}
\end{equation}
where $d_{j^v_k,q}$ is the transmission delay for $q$th packet requested in cell $k$ by user $j$ for operator $v$. Basically, constraint (\ref{eq:delvier_const}) ensures that entire data, in terms of number of bits, is delivered and constraint (\ref{eq:maxdelay_const}) implies that that transmission latency should not violate the $D^v_{max}$ which was guaranteed by the MVNO.

Given the fixed number of PRBs at each $T_s$, and the stochastic and non-stationary nature of users' requests in each cell, it would be challenging to satisfy all users' demands with a $\% 100$ guarantee unless each MVNO requests extra PRBs at each network slicing time slot, which is not cost-efficient. Therefore, our goal here is to design a controller that guarantees a probabilistic upper-bound on the user delay requirement, i.e., $Pr(d_{j^v_k,q} > D^v_{max}) < {\epsilon}_v$, with the minimum possible PRBs to satisfy such requirement.

Mathematically speaking, the sequential decision making problem that each MVNO has to solve can be formulated as follows:

\begin{equation}
\min \lim_{T\to\infty} \frac{1}{T} \sum^{T}_{T_s=1} N^v_{T_s} =\mathop{\mathbb{E}}\big\{N^v_{T_s}\big\}\\ \label{eq:mainprob}
\end{equation}

\begin{equation}
\textrm{s.t.} \quad  Pr(d_{j^v_k,q}> D^v_{max}) < {\epsilon}_v \quad    \forall k,j,q,v \\ \label{eq:const}
\end{equation}
where $N^v_{T_s}$ is the number of PRBs used in time slot $T_s$ for $v$th MVNO's clients which is equal to:  
\begin{equation}
N^v_{T_s} = \sum_{t=t_0} ^{t_0+T_s} \sum_{j}\sum_{q} l^t_{j^v_k,q}
\end{equation}

So the goal is to minimize the expected number of PRBs used over an infinite horizons of network slicing time slots, while guaranteeing that dissatisfactions of users occurs with less than ${\epsilon}_v$ probability as reflected in (\ref{eq:const}).

\section{Deep Reinforcement Learning} \label{sec:drl}
In this section, we will discuss the design of the MVNO's slice controller and how to represent the controller's logic as a reinforcement learning agent.
In a reinforcement learning problem, the agent iteratively interacts with
the environment. This environment is usually described by a
Markov decision process (MDP) which can be defined with 5-tuple $\langle \mathcal{S}, \mathcal{A},\mathcal{P}, \mathcal{R}, \gamma \rangle$	, which is 
state space $ \mathcal{S}$, action space $ \mathcal{A}$, state transition probability $Pr(s_{t+1}
|s_t, a_t) \in \mathcal{P}$, reward function $R$, and discount factor $\gamma \in \big[0, 1\big)$.
According to this notation, at each time $t$, the agent based on its current state $s_t$ takes an
action $a_t \in \mathcal{A}$, and goes from state $s_t$ to a new state $s_{t+1}$ with probability $Pr \big(s_{t+1}|s_t, a_t \big)$ and receives reward of $r_{t+1}$ from the environment. If we define
policy $\pi\big(s, a\big)$ as the probability of taking action $a_t = a$ in
state $s_t = s$, i.e, $\pi\big(s, a\big) = Pr\big(a_t = a|s_t = s\big)$, the goal
of the agent is to learn a policy that maximizes the expected
sum of discounted rewards it receives over the long run. This
sum of discounted rewards is called 'return' and is defined as
$R_t = \sum_{ i=0 }^{T} \gamma^{i+t} r_{t+i}$ where T is the length of the time horizon. We define the  optimal policy $\pi^*$,  that chooses best actions $a^*$ in  state $s$, which maximizes the  expected return without knowledge of the function form of the reward and the state transitions.

\begin{equation}
\pi^* = \argmax_{\pi} {E}_{\pi}\{ R|s_0=s \}
\end{equation}

In order to solve the optimization problem (\ref{eq:mainprob}) using DRL, first we need to define the problem in a RL setting. To this end, we need to define the states, actions and rewards for the MVNOs which are the RL agents in our context.

\subsection{States}
Once the number of PRBs has been determined, the RAN's scheduling algorithm allocates the available PRBs to incoming requests on the $T_b$ time scale. We consider various network statistics as input state to our DRL model. These features include the percentage of requests that meet the delay deadline, average delay, standard deviation of all requests in the current network slicing time slot, average and standard deviation of SNR at the client side (average PRB capacity) which takes into account the wireless propagation environment and the mobility of users, average demand-to-capacity ratio (i.e., average packet size request divided by capacity), and the number of allocated PRBs in the previous time slot. To track network behavior and user traffic patterns, these features are considered over a history window of length  $h$, rather than focusing solely on current time slot values. This approach builds a stronger model capable of capturing the temporal correlation of these features.



\subsection{Actions}
In our setting, actions determine the proper number of PRBs given the current state. This could be represented by a finite integer set, such as $\mathcal{A}=\big\{10,20,30, ...., 120\big\}$, assuming the maximum available PRBs for the entire RAN are 120. However, a more reasonable approach, which we have chosen for our setting, involves using differential actions due to the strong temporal correlation observed in PRB utilization in real-world environments. This approach prompts the DRL agent to decide on the increase or decrease in the number of PRBs for the next network slicing time slot. Thus, a possible action space in this scenario could be:

\begin{equation}
\mathcal{A}=\big\{-2^{J},-2^{J-1},...,-1,0,1,...,2^{J-1},2^{J}\big\}
\label{eq:action_set}
\end{equation} 
where the $|\mathcal{A}|=2J+1$

\subsection{Reward}

Optimal reward function could based be derived based on the
optimization problem (\ref{eq:mainprob}) and constraint (\ref{eq:const}) as below:

{\small
\begin{equation}
J_{\pi_{\theta}} = L_{\theta,\lambda} = -  \mathbb{E}\big\{  N_{T_s} \big\}+ \lambda ( Pr(d_{q}< D^{max})-(1-\epsilon)) \label{eq:reward_theory2}
\end{equation}
}

As shown Appendix  \ref{sec:reward_design}, maximizing the average reward $J_{\pi_{\theta}}$ with respect to $\theta$ is equivalent to computing the Lagrangian dual function associated with problem (\ref{eq:mainprob}).

\section{Practical Modification of Reward Function} \label{sec:pract_reward}

We can derive an optimization-based reward function as shown in previous section and Appendix  \ref{sec:reward_design}.  However, one problem that exists is that, in theory, it is assumed that the second term in the reward equation (\ref{eq:reward_theory2}) always increases with respect to $N_{Ts}$, as we expect that by increasing the number of PRBs, the probability of violating the delay constraint decreases. 

\begin{figure}
  \begin{subfigure}{\linewidth}
    \centering
    \includegraphics[width=0.7\linewidth]{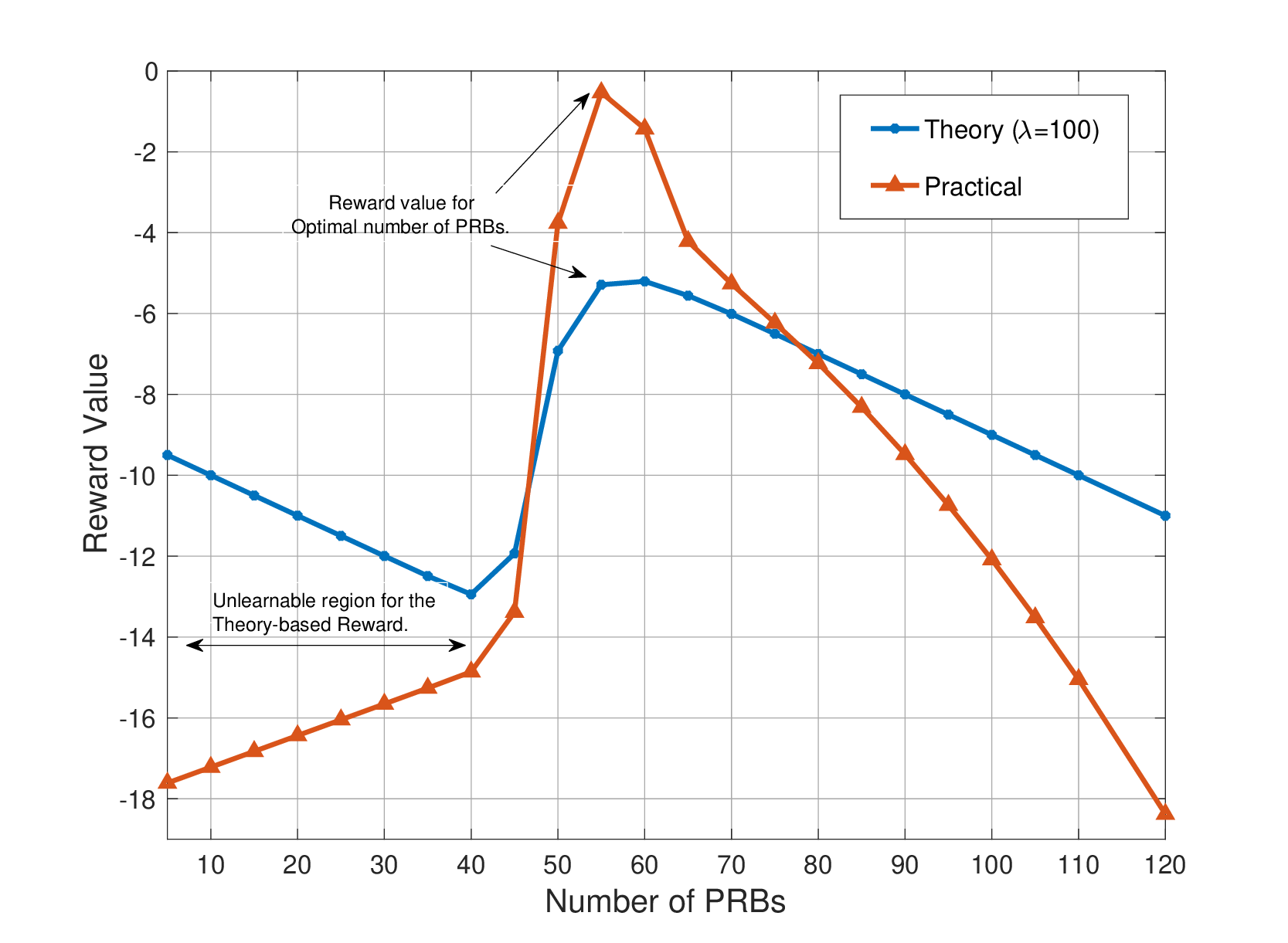}
    \label{fig:theoryVSpractice}
  \end{subfigure}
  \begin{subfigure}{\linewidth}
    \centering
    \includegraphics[width=0.7\linewidth]{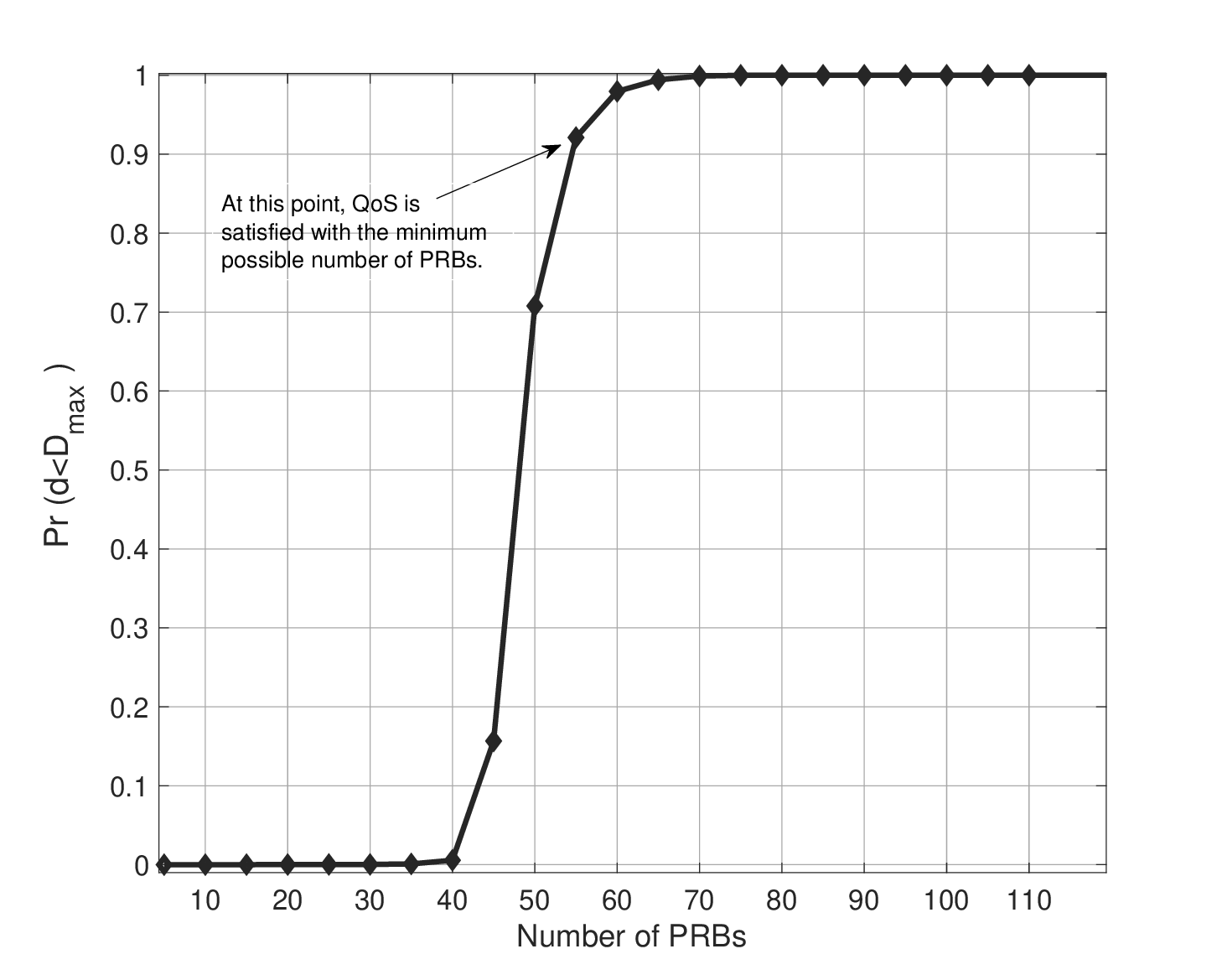}
    \label{fig:optmial_reward_delay}
  \end{subfigure}
  \caption{(a) Theory vs practical reward functions. (b) Probability of satisfying of the maximum delay constraint.}
  \label{fig:alltheoryVSpractice}
\end{figure}

However, in reality as we see in Fig. \ref{fig:alltheoryVSpractice}a, there is a threshold below which none of the packets satisfy the delay constraint. In practice, what occurs is that increasing the PRBs decreases the average delay; however, there might still be instances where none of the packet transmission delays fall below $D^{max}$, leading to a zero probability of satisfaction. Consequently, the only dominant part of the reward function, $- N_{Ts}$, would motivate the agent to decrease the requested PRBs and this would make the agent training very challenging and almost impossible. We have marked this region in Fig. \ref{fig:alltheoryVSpractice}a as an "unlearnable region" because training an agent using gradient descent-based algorithms with initial conditions in this region would fail, as the agent would eventually request zero PRBs.

Therefore, in order to have a practical reward function which could be used in different environment conditions and also provide a more robust and faster learning convergence, we propose the below modified reward function:

\[ r =
  \begin{cases}
    - \Delta \gamma_p+e^{\zeta_p \Delta + \nu_p} N^2_{T_s}      & \quad \text{if } \Delta \geq 0\\
    \Delta \gamma_n + e^{\zeta_n \Delta + \nu_n} N_{T_s}   & \quad \text{if } \Delta <0
  \end{cases}
\]
where 
\begin{equation}
 \Delta =  Pr(d_{q}< D^{max})-(1-\epsilon)
\end{equation}

This new reward function is composed of two sections. When the constraint is not satisfied ($\Delta <0$), the first term of the reward function is similar to the second term of equation (\ref{eq:reward_theory}) as an increase in the probability of satisfaction contributes linearly to the increase in the reward value, while the second term is linear in $N_{T_s}$ with an exponential coefficient that depends on $\Delta$ itself and the coefficients $\zeta_n$ and $\nu_n$. The $\zeta_n$ and $\nu_n$ can be set in a way that, when $\Delta$ is negative, the exponential term is large and motivates the agent to increase $N_{T_s}$ faster. As $\Delta $ approaches zero, the exponential term fades, and the linear term becomes dominant. Similar analysis holds for the case when $\Delta >0$.

As observed in Fig. \ref{fig:alltheoryVSpractice}a, both the original and modified reward functions reach their maximum at the same point (PRB=55), where the probabilistic constraint is satisfied with the minimum number of PRBs (as shown in Figure \ref{fig:alltheoryVSpractice}b for $\epsilon=0.1$). Therefore, the modified reward function achieves the same optimal point while being shaped in a way that promotes more robust and faster agent training. It is important to note that, for convergence guarantee, we clip the output of the function to be bounded between $(-R_{max},0)$.

\vspace{-2.5mm}

\section{DRL Personalization}\label{sec:PFDR}
After the training phase, each agent associated with a particular MVNO in a cell would develop its own DRL model. However, we want to further enhance the performance of these models by leveraging the knowledge and expertise of other agents. This is where personalized federated learning comes into play. Personalized federated learning \cite{huang2021personalized} offers a key benefit of creating highly personalized models for each agent without compromising data privacy.  By combining the knowledge and models from multiple agents, we can generate a more robust and accurate model tailored for a specific MVNO and RAN.

Mathematically speaking, considering $N$ DRL agents with a model weights vector $\mathbf{W}=\left[W_0, W_1, ..., W_{N-1}\right]$, each agent $i$ aims to find the optimal aggregation coefficient vector $\boldsymbol{\alpha_i}$. This vector is designed to maximize the performance of the resulting personalized model $W^p_i$, as described in (\ref{eq: personalized_weight}), when executed in its own environment $i$.

\begin{equation}\label{eq: personalized_weight}
  W^p_i=\sum_{j=1}^N  \alpha_{i,j} W_j
\end{equation}

Here, we propose a reward-based personalization for DRL models. The main idea behind this method is that each agent utilizes the model weights of other agents based on their respective performance. Additionally, we will compare this approach with other model aggregation methods such as those considering similarities in model weight \cite{nagib2023accelerating} and traffic patterns and wireless environment features \cite{rezazadeh2022specialization} and also model averaging \cite{tehrani2021federated}.

\subsection{Personalizing Based on Feature Weights}
In this method, each agent would aggregate other agents' models based on a feature vector that represent the wireless environment and QoS of the associated RAN and MVNO respectively. We define the feature vector of the $i$th agent model as $ \mathbf{f_i}=\left[D_{{max}_i},\epsilon_i,\tilde{K_i},\tilde{C_i},\tilde{T_i}\right] $, 
where $D_{{max}_i}$ and $\epsilon_i$ are related to QoS of $i$th agent, $\tilde{K_i}$, $\tilde{C_{i}}$ and $\tilde{T_i}$ represent the average number of users, average channel capacity and average number of requested data packets for the $i$th environment respectively. In order to compute distances of two feature vectors, we use the following distance function:

\begin{equation}\label{eq:fed}
  dist(\mathbf{f_i},\mathbf{f_j})=\sum_{m} \exp{\frac{\|	f^m_i-f^m_j \|^2}{\sigma_m}}
\end{equation}
where $\sigma_m$ is a temperature parameter and could be proportional to the standard deviation of $m$th feature. The aggregation coefficient $\alpha_{i,j}$ can be obtained as follows: 

\begin{equation}\label{eq:fed}
  \alpha_{i,j}=\frac{dist(\mathbf{f_i},\mathbf{f_j})}{\sum_{j=1}^N dist(\mathbf{f_i},\mathbf{f_j})}
\end{equation}

\subsection{Personalizing Based on Model Weights}
In this method, each agent would obtain the aggregation coefficient based on the similarity of the DRL model weights. Here we use $L_2$ norm as distance function in this approach:

\begin{equation}\label{eq:fed}
  \alpha_{i,j} = \frac{\|W_i-W_j \|^2}{\sum_{j=1}^N \|W_i-W_j \|^2}
\end{equation}

\subsection{Reward Based Personalization}
In this method, the $i$th agent computes the aggregation coefficients for other agents' models based on their performance in its own environment. Agent $i$ tests model ${W_j}$ on its own environment $i$ for $T$ episodes and obtains the average reward $\hat{R}_{i,j}^T$. The aggregation coefficient is then obtained as follows:

\begin{equation}\label{eq:reward Person}
  \alpha_{i,j} =\frac{ e^{\beta \hat{R}_{i,j}^T } }{ \sum_{j} e^{ \beta \hat{R}_{i,j}^T  }} 
\end{equation}
where $\beta$ is a temperature parameter, the algorithm behaves similarly to federated averaging when $\beta$ is close to zero. As $\beta$ increases, the algorithm prioritizes selecting the best model.

In Appendix \ref{sec:oran_comp} we discussed how the training and personalization of different models for different RANs would be aligned in O-RAN compliance architecture.

\begin{table*}[tb] 
\begin{center}
\caption{Performance comparison of different policies.}\label{tab:comparison} 
\begin{tabular}{ |p{3.5cm}|p{1.6cm}|p{1.6cm}|p{1.6cm}|p{1.6cm}|p{1.6cm}|}
 \hline
 Policy & PDA-DRL & MD-DRL & Heuristic & Fixed-Av & Fixed-Max\\
 \hline
 Average PRB utilization &  56.19  &  54.81 & 54.68  &   60.0 &   120.0  \\
 \hline
 $Pr(d_{q}< D^{max})$ & 0.89 & 0.75 & 0.78  &   0.72 &   1.0  \\
 \hline
   Mean delay (ms) & 3.01 & 4.86 & 4.47  &    6.03 &   1.11  \\
 \hline
    Delay STD (ms) & 1.65 & 2.48 & 2.26  &    1.60 &   0.26  \\
 \hline
   Average reward & -1.69 &  -4.39 &  -3.45  &    -5.01 &   -18.37  \\
 \hline
\end{tabular} 
\end{center}
\end{table*}

\section{Simulation Results} \label{sec:simulation}

\begin{figure}
\begin{subfigure}{.5\textwidth}
  \centering
  \includegraphics[width=.65\columnwidth]{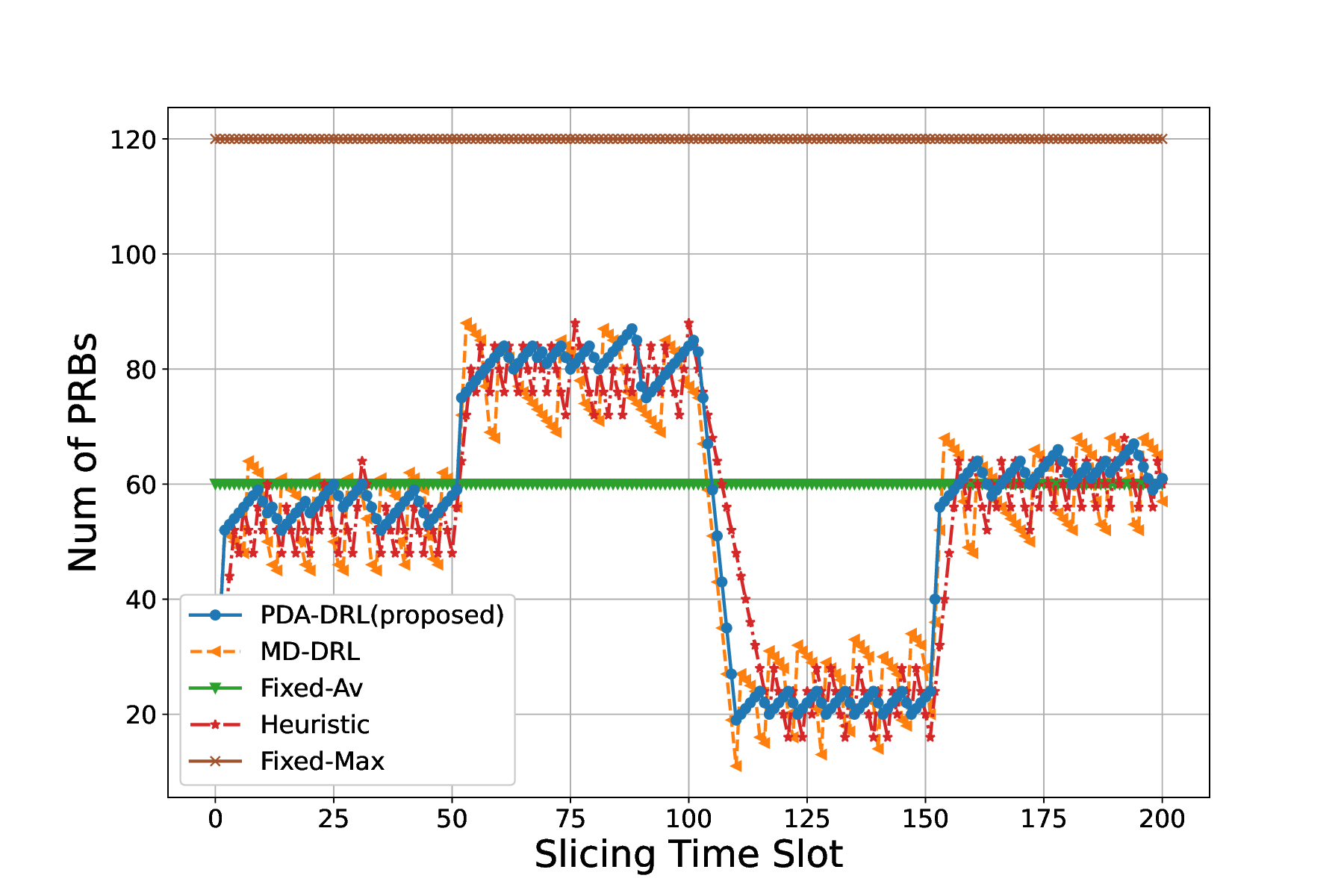}
  \caption{PRB usage comparison between different policies.}
  \label{fig:sub-first}
\end{subfigure}
\begin{subfigure}{.5\textwidth}
  \centering
  \includegraphics[width=.65\columnwidth]{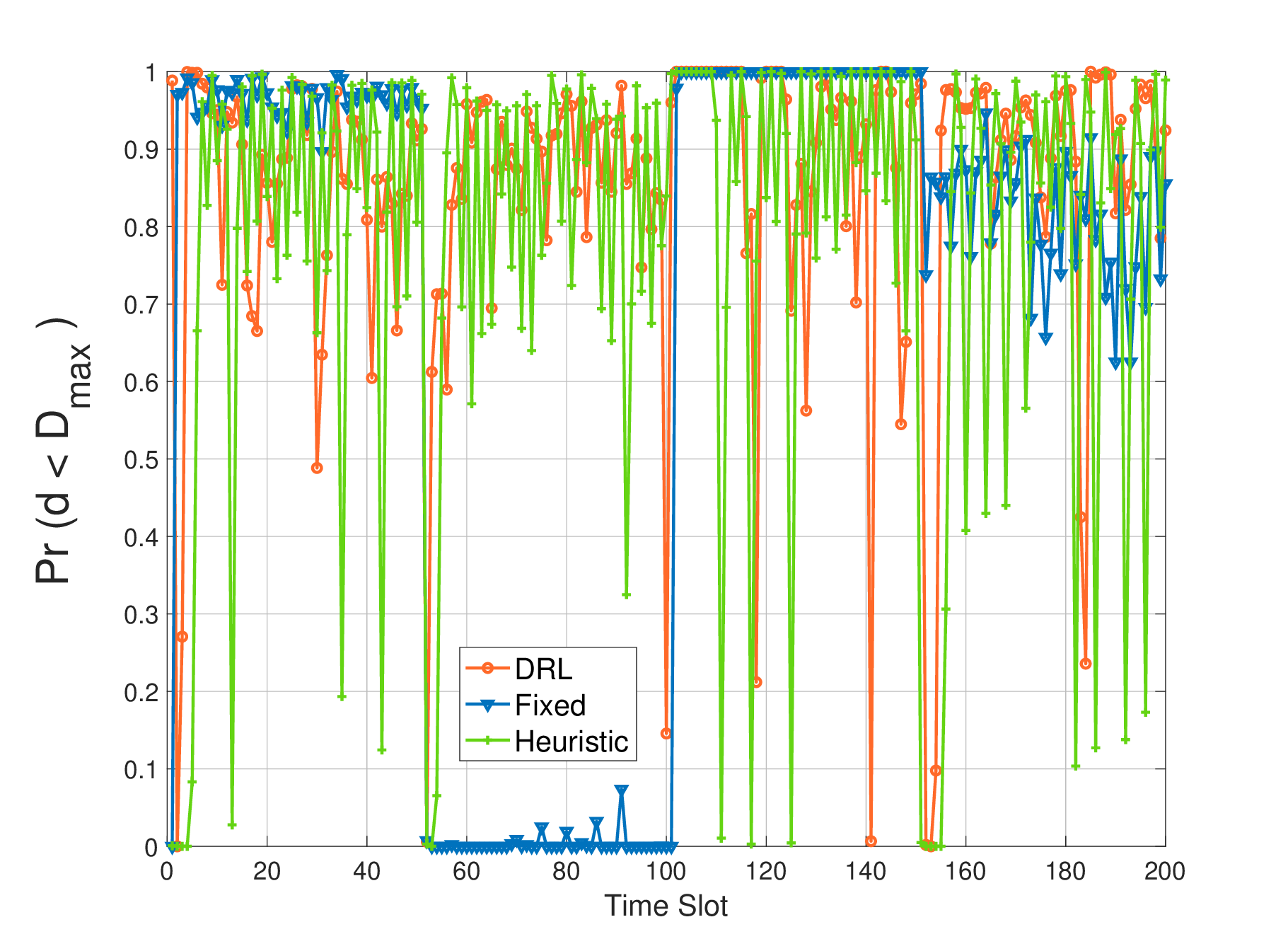}
  \caption{QoS satisfaction comparison between different policies.}
  \label{fig:sub-second}
\end{subfigure}
\begin{subfigure}{.5\textwidth}
  \centering
  \includegraphics[width=.65\columnwidth]{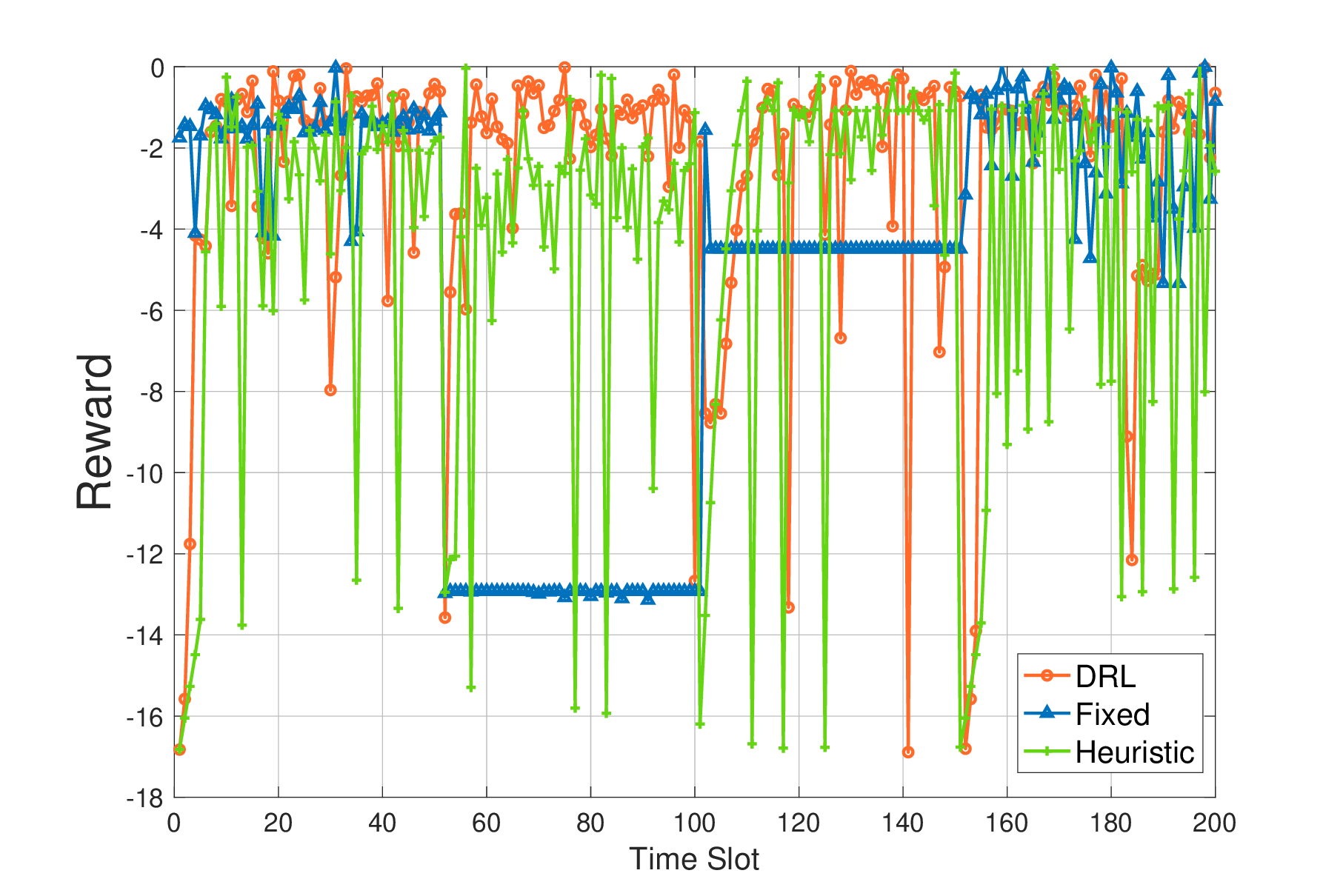}
  \caption{Reward comparison between different policies.}
  \label{fig:sub-second}
\end{subfigure}
\caption{Performance comparison of DRL-based methods, heuristic and fixed polices when there is a dynamic user traffic with increasing and decreasing patterns.}
\label{fig:policy comparison}
\end{figure}

In this section, we define the simulation scenarios, specify the parameters, and provide a comprehensive performance evaluation of our proposed algorithms. The DRL models have been implemented in PyTorch, and we used the Adam optimizer with a learning rate of $lr=0.0003$ for training. Our chosen DRL algorithm is deep policy gradient, and the model consists of two hidden layers with 128 and 64 neurons, respectively, followed by ReLU activation functions. The output dimension is set to be the same as the number of actions, we set $J=5$ in  (\ref{eq:action_set}).

In the simulation scenario, we set $T_b=1$ ms, which is the duration of an LTE subframe. We also set $T_s=1$ s, or equivalently $H=1000$, meaning the agent makes decisions every 1 second, therefore, the number of PRBs remains fixed for the subsequent $1000$ time slots. We consider different QoS requirements for MVNOs in terms of maximum transmission latency. Two different services with maximum transmission latencies of $5$ ms and $10$ ms are considered. For the probabilistic constraint, we examine two different epsilon values: $\epsilon=0.1$ and $\epsilon=0.3$.

To simulate mobility and time-varying channel conditions, we introduce a Doppler frequency range of $5$ Hz to $50$ Hz. The number of users per slice ranges from $2$ to $6$. Additionally, for each user, we assume the data request packet sizes can follow three different distributions: small, medium, and large. Finally, we assume a total of $150$ available PRBs in each cell.

\begin{figure*}[t]
\centering
\begin{subfigure}{.24\textwidth}
  \centering
  \includegraphics[width=\linewidth]{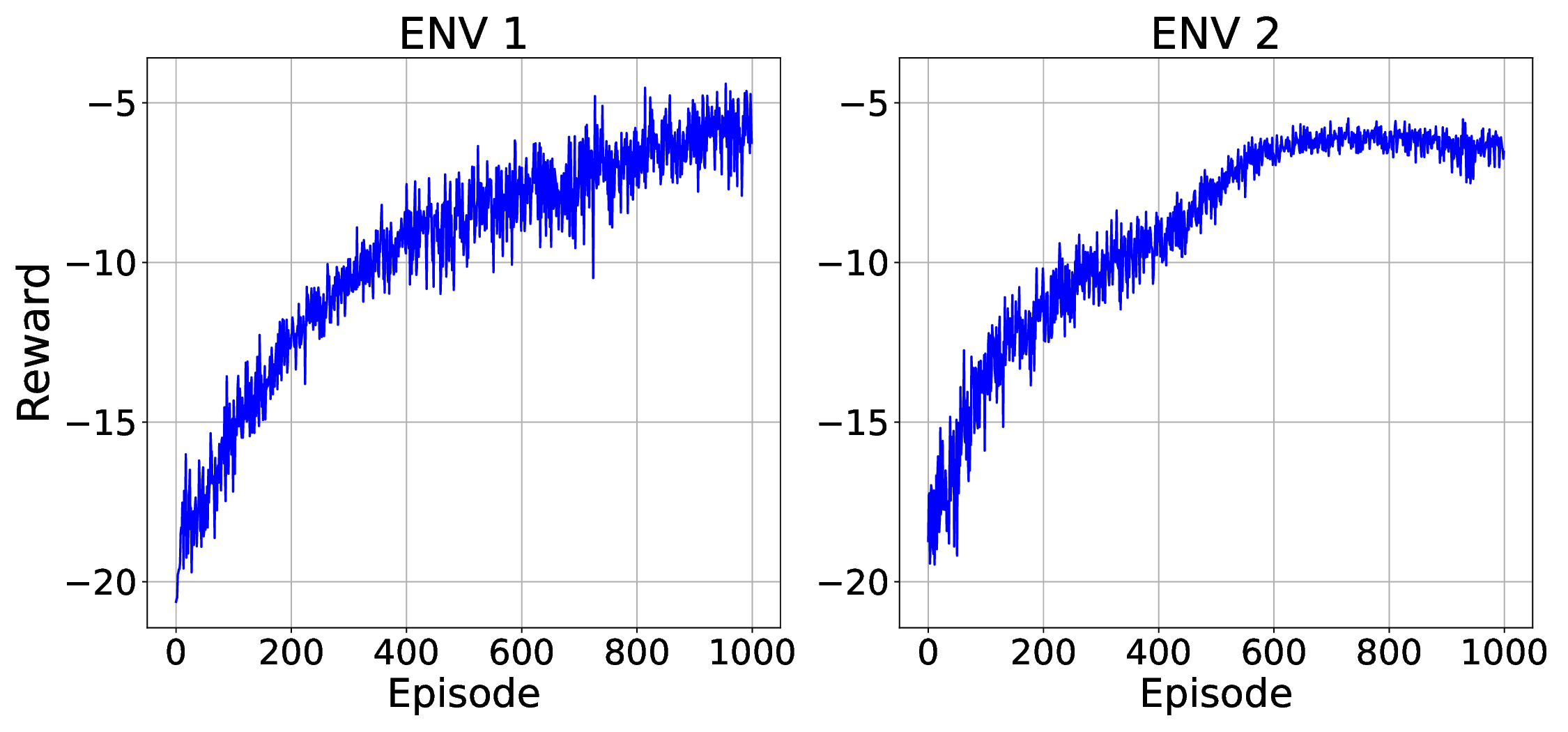}
  \caption{Reward.}
  \label{fig:sub-first}
\end{subfigure}%
\begin{subfigure}{.24\textwidth}
  \centering
  \includegraphics[width=\linewidth]{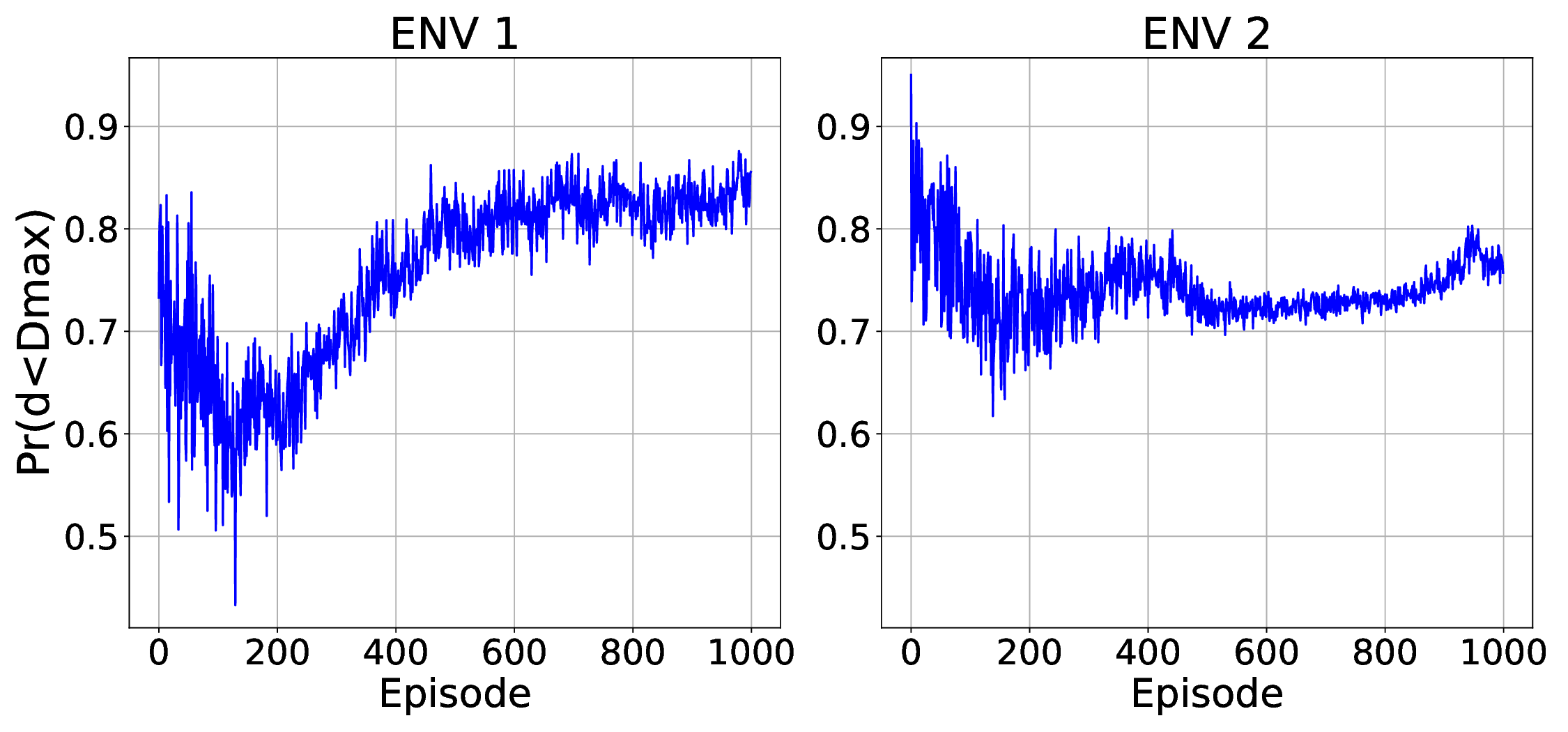}
  \caption{Probabilistic constraint.}
  \label{fig:sub-second}
\end{subfigure}%
\begin{subfigure}{.24\textwidth}
  \centering
  \includegraphics[width=\linewidth]{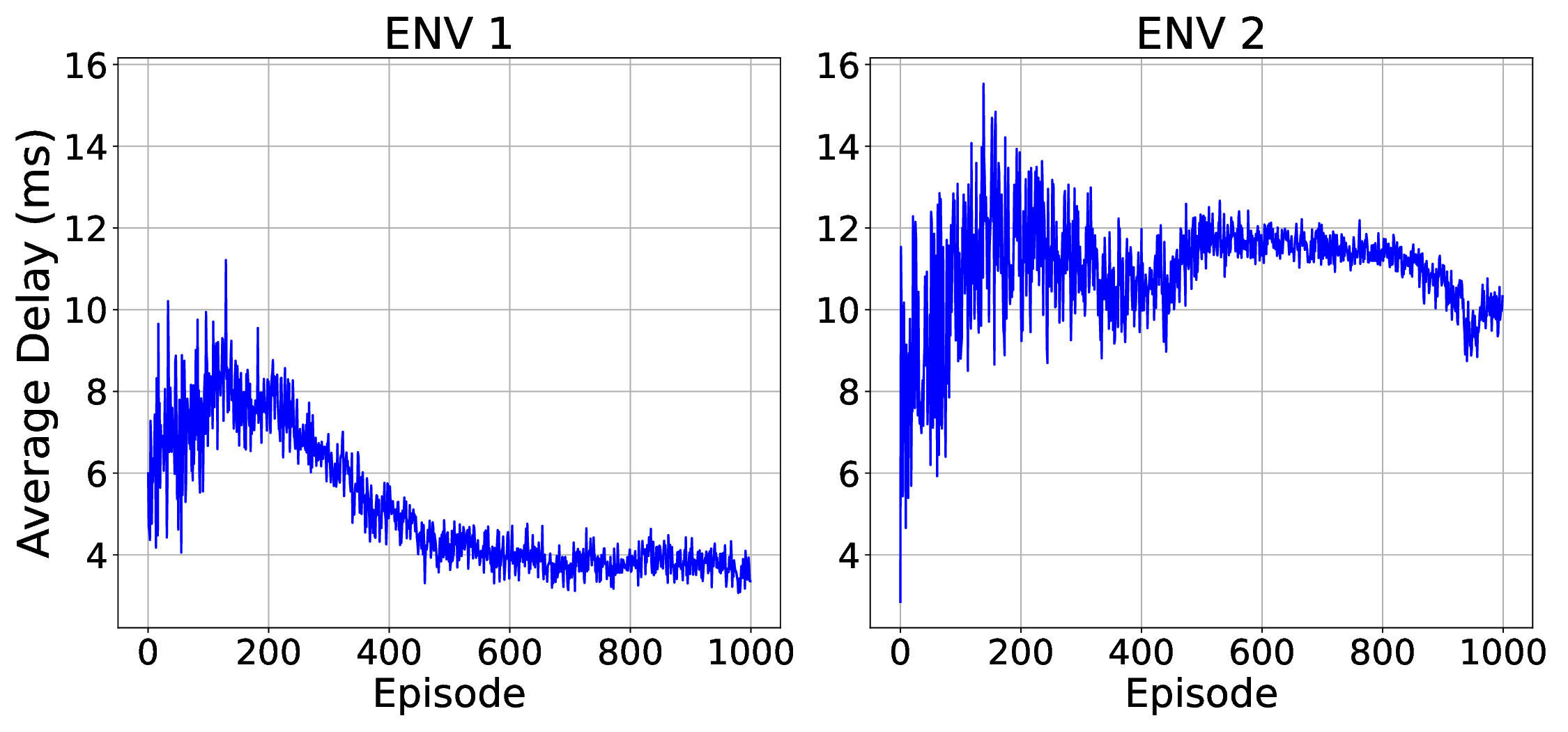}
  \caption{Average delay.}
  \label{fig:sub-third}
\end{subfigure}%
\begin{subfigure}{.24\textwidth}
  \centering
  \includegraphics[width=\linewidth]{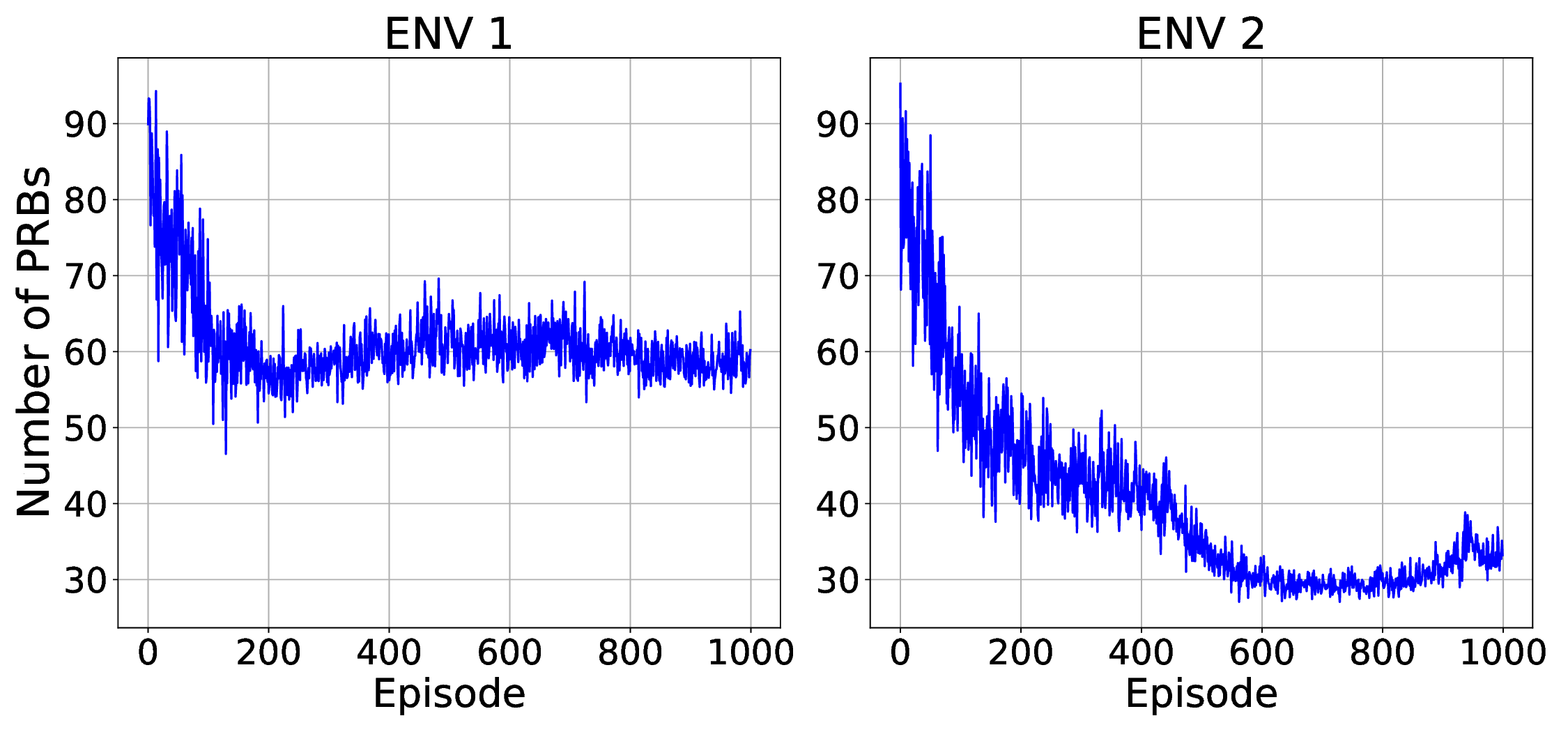}
  \caption{ Number of PRBs.}
  \label{fig:sub-fourth}
\end{subfigure}
\caption{Convergence plots of different features in two distinct environments with different QoS requirements.}
\label{fig:convergence}
\end{figure*}

Fig. \ref{fig:convergence} shows the convergence plots of the DRL agent during training in two different environments. In ENV1, we have $D^{max} = 5$ ms and $\epsilon=0.1$, while in ENV2, we have $D^{max} = 10$ ms and $\epsilon=0.3$. As we can see, the average delay goes below the $D^{max}$ in both cases, while the satisfaction probability getting close to $1-\epsilon$ as the model converges. Based on the wireless environment and QoS requirement, the agent is converging to the optimal number of PRBs, which for ENV1 is somewhere around 60 and for ENV2 is close to 30. It can be observed that the proposed reward function can appropriately balances the excessive use of PRBs and maximum delay violation while providing a smooth convergence.

In Figure \ref{fig:policy comparison}, we compare our proposed solution PDA-DRL, with four other baselines: Mean Delay DRL (MD-DRL), Fixed-Av (Average Fixed), Fixed-Max (Maximum Fixed), and the heuristic policy, within a simulation scenario set at $D^{max} = 5$ ms and $\epsilon = 0.1$. For MD-DRL, the DRL model is trained with a reward function that maximizes when the average transmission delay is $5ms$. This criterion was selected to establish a DRL baseline and consider the average performance, as is common in many state-of-the-art solutions. The Fixed-Av policy employs a predetermined number of PRBs based on the average traffic demand of the slice, whereas the Fixed-Max policy uses a sufficient number of PRBs to ensure a 100\%  probability of meeting the delay constraint during peak traffic. This baseline assesses the excessive PRB usage required for an MVNO to perfectly satisfy QoS constraints with a fixed approach. Conversely, the heuristic policy dynamically adjusts the number of PRBs in response to the network's immediate state, increasing PRBs when constraint (\ref{eq:const}) is not met and reducing them otherwise, to avoid excess usage.

As observed in Fig. \ref{fig:policy comparison}, the fixed policies struggle to dynamically adapt to fluctuating traffic, leading to a trade-off between excessive PRB usage and QoS degradation during varying traffic demands. While Fixed-Max ensures a 100\% guarantee of transmission delay staying below 5 ms, it consumes more than twice the number of PRBs compared to the other baselines. The MD-DRL, as expected, maintains an average delay below 5 ms, but the delay standard deviation (STD) indicates that still a considerable portion of packets experience transmission delay longer than 5 ms. The heuristic policy, which adjusts PRBs based on current satisfaction probabilities, tends to fluctuate between satisfied and unsatisfied QoS states. Although it performs better than the Fixed  baselines and MD-DRL, it still falls more than 12\% short of the QoS constraint. On the other hand, the PDA-DRL uses almost the same number of PRBs as the heuristic and MD-DRL approaches while closely meeting the QoS constraint within a 1\% margin. \textbf{Notably, the PDA-DRL approach demonstrates superior performance by achieving a 38\% reduction in average delay and 33\% reduction in STD compared to MD-DRL. This illustrates that a percentile-based formulation can also provide a tighter bound on average delay.} The summarized performance metrics of these methods are reported in Table \ref{tab:comparison}.

\begin{figure}
\centering
  \includegraphics[width=.99\columnwidth]{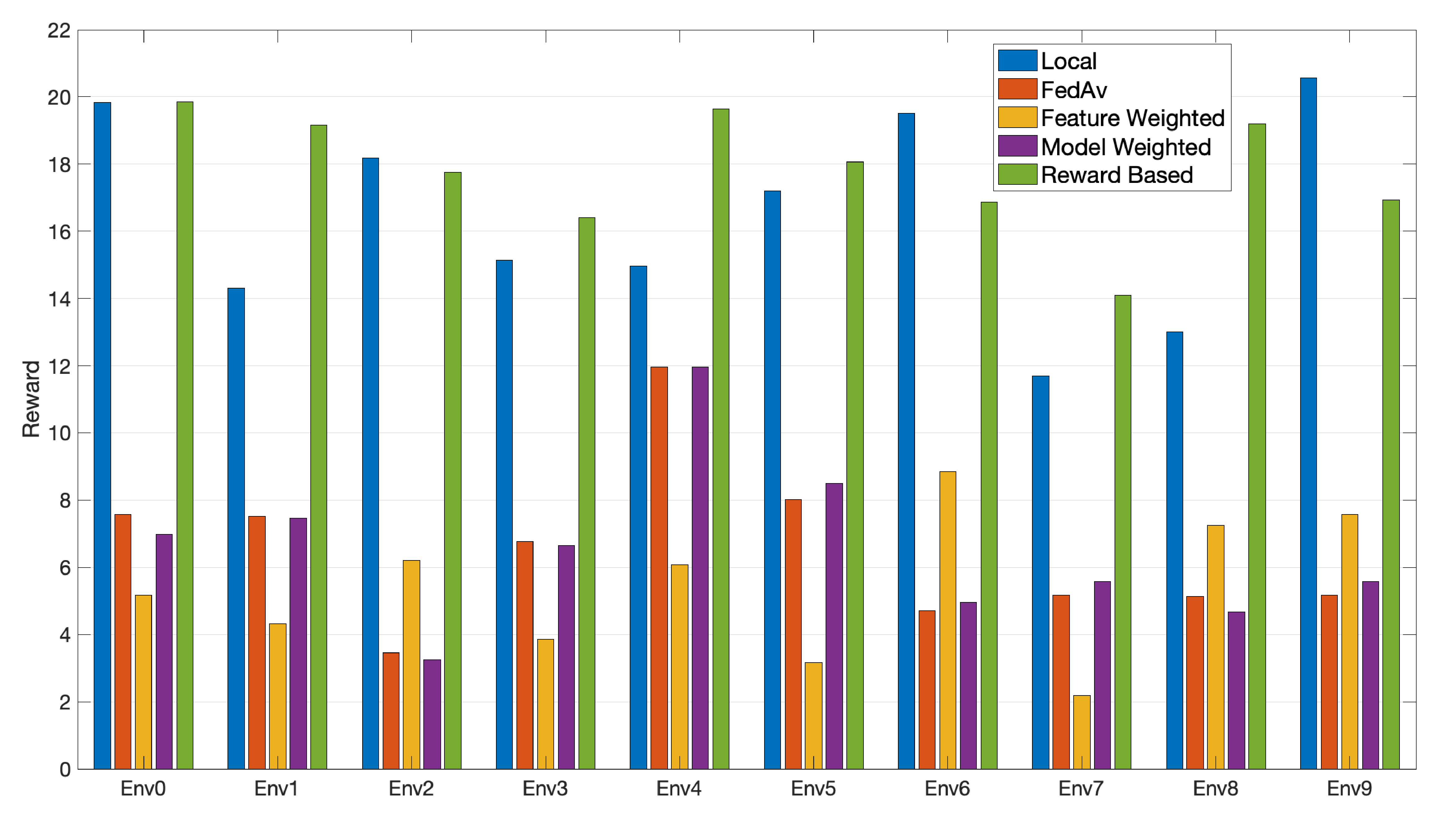}
  \caption{\small Comparison between different aggregation methods.}
  \label{fig:aggregation comparison}
\end{figure}

In Fig. \ref{fig:aggregation comparison}, we present a comparison of various personalization methods' performance across 10 distinct environments. These environments are characterized by significant differences in QoS requirements ($\epsilon$, $D^{max}$) and wireless conditions, including average CQI, Doppler frequency, number of users, and types of traffic. As expected, methods like FedAV, which rely on naively averaging model weights, show a poor performance. Similarly, aggregation based on model weight and feature similarity does not significantly enhance performance. In contrast, reward-based personalization demonstrates superior performance, outstripping all other aggregation methods by a considerable margin and rivaling the performance of local models. It is worth noting that, to facilitate comparison in Fig. \ref{fig:aggregation comparison}, we introduced a positive bias to the obtained rewards, as they are inherently negative. For the aggregation coefficient in (\ref{eq:reward Person}), we set $\beta = 3$ and $T = 10$.

In Appendix \ref{sec:exp framework}, we have validated our DRL solution through experimental framework which is built on top of  Colosseum \cite{bonati2021colosseum} and Scope\cite{bonati2021scope}.

\section{Conclusion} \label{sec:con}

We proposed PDA-DRL for delay-aware RAN slicing in O-RAN, optimizing PRB allocation via LLN-based rewards and reward shaping. Our solution outperforms baselines (e.g., MD-DRL) and introduces collaborative DRL personalization, excelling in diverse QoS scenarios.

\bibliography{ref}
\bibliographystyle{icml2025}

\newpage
\appendix
\onecolumn

\section{Reward Function Design} \label{sec:reward_design}
To devise a suitable reward function, it is essential to consider the progress of the $v$th MVNO, acting as a DRL agent, towards achieving the objectives of the sequential optimization problem (\ref{eq:mainprob}), while also ensuring compliance with the constraint (\ref{eq:const}). Consequently, the reward function should focuses on tracking packets that meet the transmission delay deadline and imposes penalties for excessive PRB utilization. 
 Following a similar approach to that used in \cite{raeis2021queue} for RL reward design, for each packet $f_{j,k,q}$ we define variable $z_{q}$ as follows:

\[ z_{q} =
  \begin{cases}
    1       & \quad \text{if } \sum_{t=t_0} ^{t_0+D^{max}} C^t_{j,k}l^t_{j,k,q} \geq f^{t_0}_{j,k,q}
\\
    0  & \quad \quad \quad \quad O.W.
  \end{cases}
\]

Since our focus here is on a single DRL agent, we will drop the index $v$. For the sake of simplicity in notation, we will use the index $q$ to refer to a packet for the remainder of the analysis.

Now we can define an immediate reward, based on the delay requirement of packet $q$ as:

\[ r_{q} =
  \begin{cases}
    u_o       & \quad \text{if } z_{q}=0 \\
    u_1  & \quad \text{if } z_{q}=1
  \end{cases}
\]

\begin{equation}
r_t=r(s_t,\pi_{\theta}(s_t))=\sum_{q \in \mathcal{A}_{T_s}} \frac{r_q}{\mathbb{E}[n_a]}- N_{T_s}
\end{equation}
where $\mathcal{A}_{T_s}$ is the set of arrivals in time slot $T_s$ and $\mathbb{E}[n_a]$ is the average number of arrivals.

Now, the average reward per time step can be calculated using the defined reward function as follows:

\begin{equation}
J_{\pi_{\theta}}=\int_{\mathcal{S}} p_{\pi_{\theta}}(s)\mathbb{E}\big\{r(s,\pi_{\theta}(s))\big\}d_s
\end{equation}
where $p_{\pi_{\theta}}(s)$ is the steady state distribution of the states when following policy $\pi_{\theta}$ and is defined as follows:

\begin{equation}
p_{\pi_{\theta}}(s')=\int_{\mathcal{S}}\sum_{t=1}^{\infty} \gamma^{t-1}p(s)p       
    \big(s'|s, t,\pi_{\theta} \big) 
\end{equation}

In the above equation $p \big(s'|s, t,\pi_{\theta} \big) $ denotes the transition probability from state $s$ to state $s'$ after $t$ time step under policy $\pi_{\theta}$.

Even with a fixed state $s$ and action $a=\pi_{\theta}(s)$, the reward $r(s,a)$ can still be random due to the stochasticity arising from random packet arrivals and packet sizes in the subsequent slicing time slot. To calculate this expectation, we would have:

\begin{equation*}
\mathbb{E}\big\{r(s_t,\pi_{\theta}(s_t))|s_t=s\big\}= \mathbb{E}\big\{\sum_{q \in \mathcal{A}_{T_s}} \frac{r_q}{\mathbb{E}\big\{n_a\big\}}- N_{T_s}|s_t=s\}
\end{equation*}

defining $\mathcal{Z}$ as the set of packets in the current time slot which meet the deadline requirements: 

\begin{equation}
\mathcal{Z}=\big\{z | z_{q}=1 , \quad \forall q \big\}
\end{equation}

then the expectation would be:

\begin{align*}
& \mathbb{E}\big\{ \sum_{i \in \mathcal{Z}} \frac{u_1}{\mathbb{E}\big\{n_a\big\}} + \sum_{i \in \mathcal{Z^c}} \frac{u_0}{\mathbb{E}\big\{n_a\big\}}  - N_{T_s} |s\big\} \\
& = u_1 \frac{\mathbb{E}\big\{  |\mathcal{Z}|  \big|s \big\}}{\mathbb{E}\big\{n_a\big\}}+ u_0\frac{\mathbb{E}\big\{  |\mathcal{Z}^{\mathcal{c}}|  \big|s \big\}}{\mathbb{E}\big\{n_a\big\}} - \mathbb{E} \big\{ N_{T_s}  |s \big\}
\end{align*}

Assuming  $T_s > > T_b $, then using LLN we can approximate the previous term as:

\begin{equation*}
\approx u_1 Pr(d_{q}< D^{max}| s) + u_0 Pr(d_{q}> D^{max}| s)- \mathbb{E}\big\{ N_{T_s} | s \big\}
\end{equation*}

By choosing $u_1 = \lambda \epsilon $  and $u_0 = -\lambda (1-\epsilon)$ we would have:

\begin{equation}
J_{\pi_{\theta}} = \lambda ( Pr(d_{q}< D^{max})-(1-\epsilon))-  \mathbb{E}\big\{  N_{T_s} \big\}
\end{equation}
where $ \lambda$ specifies the trade-off between the QoS constraint and the utilized PRBs. Interestingly with this rewards design procedure, the value of  $J_{\pi_{\theta}}$ is similar to the dual form of the main optimization problem. If we consider the minimization problem (\ref{eq:mainprob}) and the constraint (\ref{eq:const}), the associated Lagrangian function would be:

\begin{equation}
L_{\theta,\lambda} = -  \mathbb{E}\big\{  N_{T_s} \big\}+ \lambda ( Pr(d_{q}< D^{max})-(1-\epsilon)) \label{eq:reward_theory}
\end{equation}

We have thus proven that maximizing the average reward $J_{\pi_{\theta}}$ with respect to $\theta$ is equivalent to computing the Lagrangian dual function associated with problem (\ref{eq:mainprob}).

\newpage

\section{ORAN Compliance} \label{sec:oran_comp}

\begin{figure}[t!]
\centering
        \includegraphics[width=.7\columnwidth]{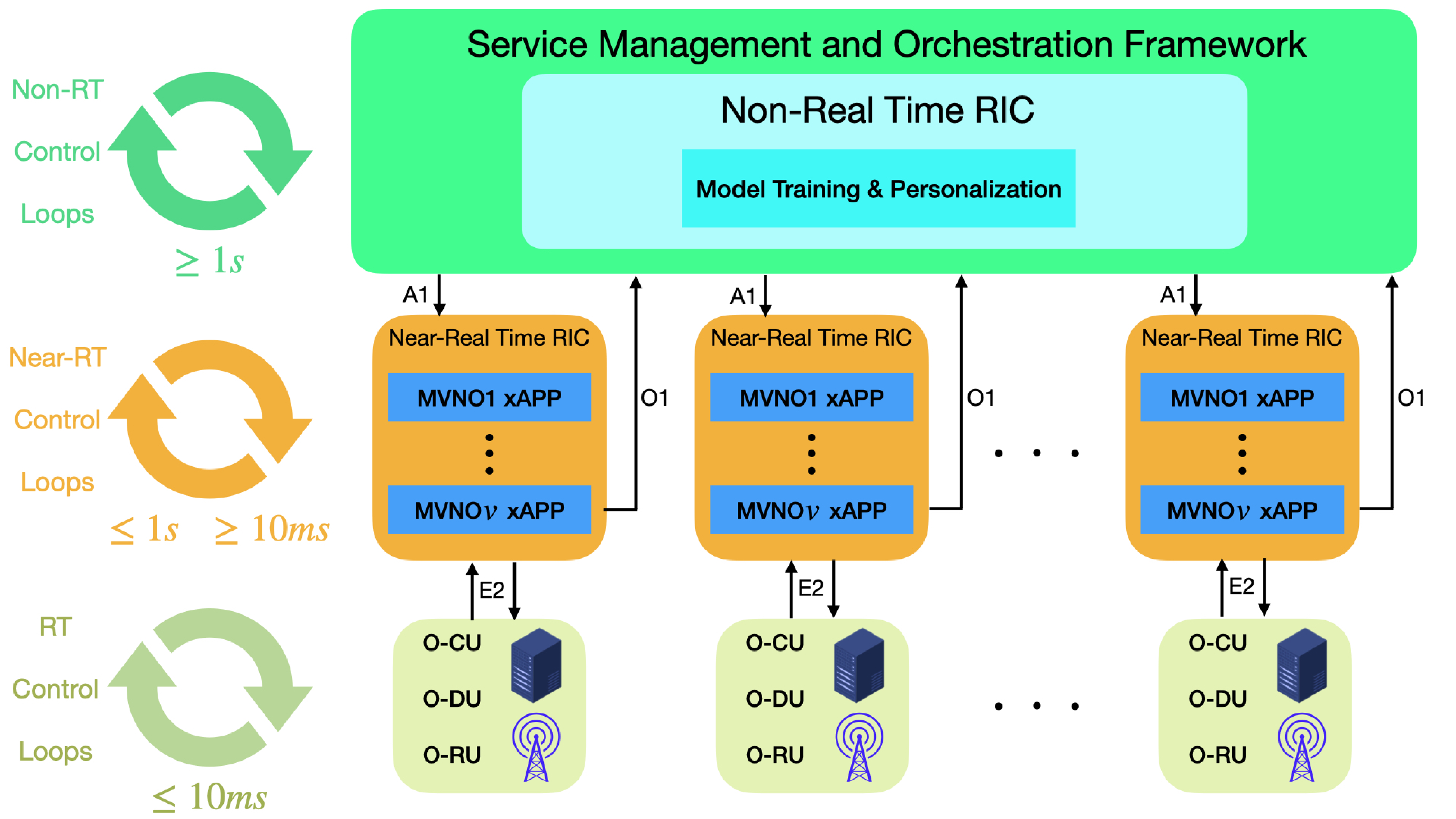}
        \caption{ORAN compliant system architecture.}
        \label{fig:oran_compl}
\end{figure}

In Fig. \ref{fig:oran_compl}, we depict the design we've crafted, which aligns with the principles of the O-RAN initiative \cite{ORANArchitectureDescription2023}. O-RAN leverages cloud RAN (C-RAN) principles and the increasingly software-defined implementations of wireless communications and networking functions. Unlike legacy interfaces that are vendor-specific and controlled by major industry players, it defines open interfaces and an open architecture to foster innovation at all layers. Central to O-RAN's architecture are the non-real-time (non-RT) RIC and the near-real-time (near-RT) RIC. The non-RT RIC is responsible for RAN optimization over broader timescales, such as seconds to minutes or longer periods of hours or days, and typically involves the training of ML models and the formulation of control policies. The model personalization discussed in Sec \ref{sec:PFDR} would occur in this layer, as it could have access to all DRL models, and model aggregation would take place on a larger timescale, like days.

The resulting models are then communicated through the A1 interface to the near-RT RICs distributed across the network. The near-RT RIC facilitates optimization and control functions within a shorter timescale, from 10 milliseconds to 1 second, and oversees monitoring of both the O-RAN Central Unit (O-CU) and the O-RAN Distributed Unit (O-DU), which host eNBs/gNBs as virtualized functions. As shown in Fig. \ref{fig:oran_compl}, each MVNO can run its own DRL-based slicing xApps in the near-RT RIC, tailored for that specific RAN region and the MVNO's QoS requirements.

Local DRL agents, or xApps, associated with the near-RT RICs, would collect data, make local decisions, and leverage the E2 interface to communicate these radio policies to the RAN. The E2 interface also serves as the conduit for transmitting RAN's KPIs back to the agents, for example, the state of the network, which includes the PRB utilization, buffer size, channel CQI, and related network features needed for ongoing model training and monitoring. This bidirectional communication facilitates a continuous feedback loop, enabling real-time adaptation and optimization of network operations.

\newpage

\section{Experimental Framework} \label{sec:exp framework}
\begin{figure}
\centering
	\begin{subfigure}[h]{.9\columnwidth}
		\centering
		\includegraphics[width=\columnwidth]{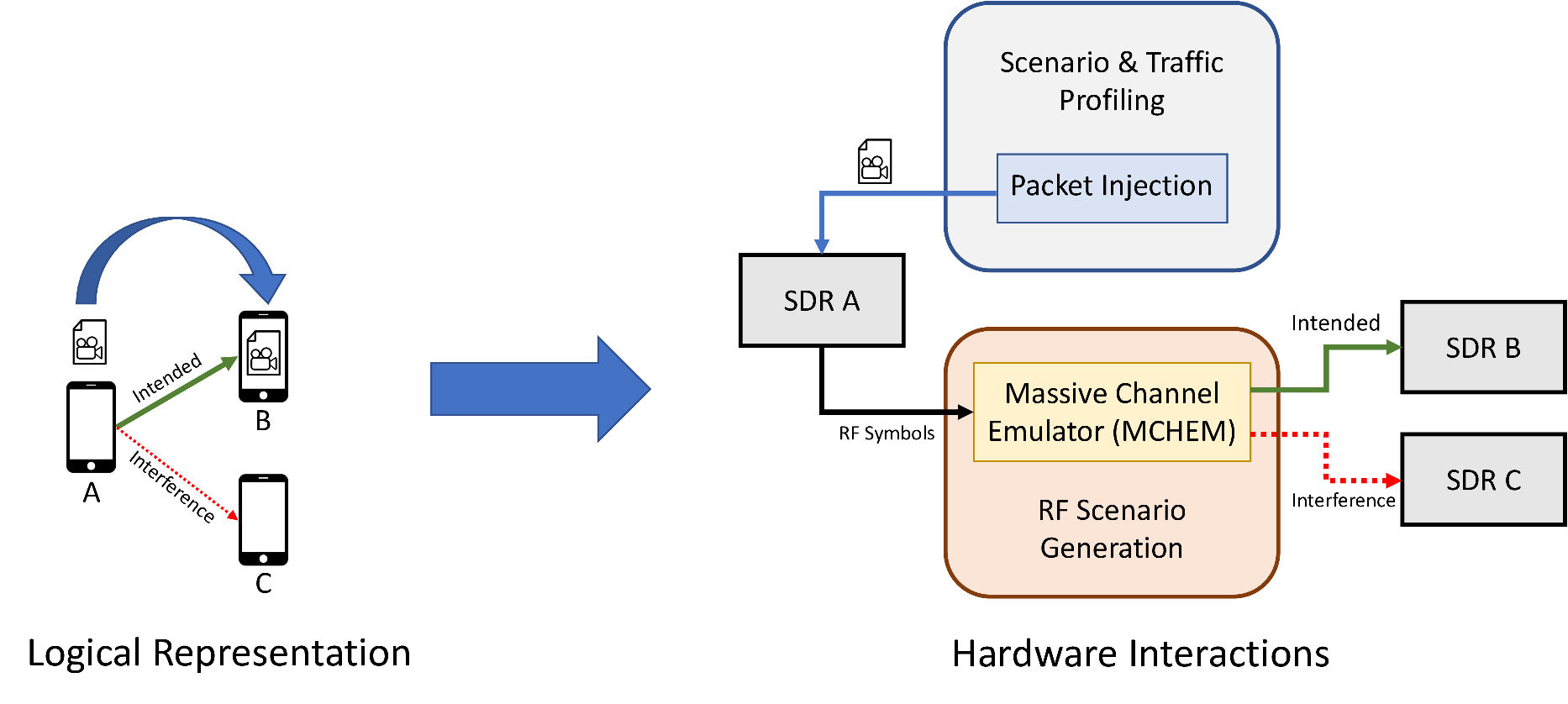}
		\caption{Implementing downlink channel on Colosseum}
		\label{fig:colosseum_downlink}
	\end{subfigure}
	\begin{subfigure}[h]{\columnwidth}
		\centering
		\includegraphics[width=.9\columnwidth]{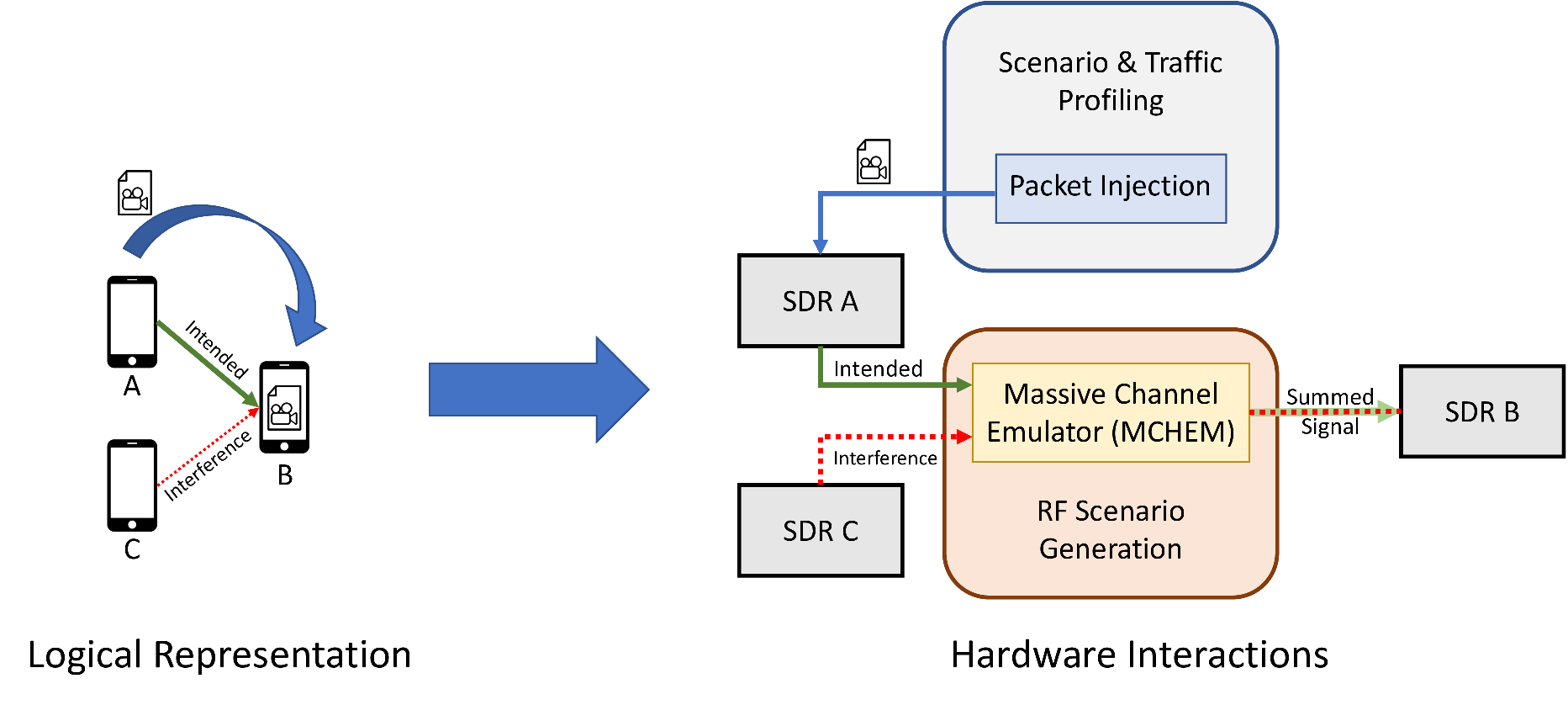}
		\caption{Implementing uplink channel on Colosseum}
		\label{fig:colosseum_uplink}
	\end{subfigure}
    \caption{Demonstration of signal transmission in Colosseum.}
\end{figure}

In this section we describe our experimental setup which utilizes  software-defined radios (SDR) in order to validate our proposed DRL model results under real-life scenarios. All of the experiments were performed on the world's largest RF emulator,  Colosseum \cite{bonati2021colosseum}, which is an extensive network of high-end servers and SDR devices. The SDR hardware used in the Colosseum is USRP X310, a high-end SDR device with a 200 MHz bandwidth per channel. The Colosseum testbed is composed of 256 SDRs connected to another 256 SDRs which enables it to output up to 65K independent RF channels. Although the testbed is built for wireless experiments, all SDRs are connected via coaxial cables. To emulate the wireless environment, every SDR-generated signal goes through a Massive Channel Emulator (MCHEM) which is composed of 512 complex-valued FIR taps that are used to apply environmental artifacts to the signals such as fading, path-loss, and interference. For example as shown in Fig. \ref{fig:colosseum_downlink}, when SDR A sends a signal towards SDR B, SDR C would receive a distorted copy of the signal as interference based on the used RF scenario (e.g. applying node mobility or multi-path effects). Similarly in Figure \ref{fig:colosseum_uplink}, when SDR A sends a signal to SDR B while having SDR C is transmitting in the background, SDR B would receive the summed signal of SDR A and SDR B transmissions. This testbed design would provide a tighter control over the RF environment that is being tested instead of radiating the RF signal into the air.

\begin{figure}[h]
    \centering
    \includegraphics[width=.8\columnwidth]{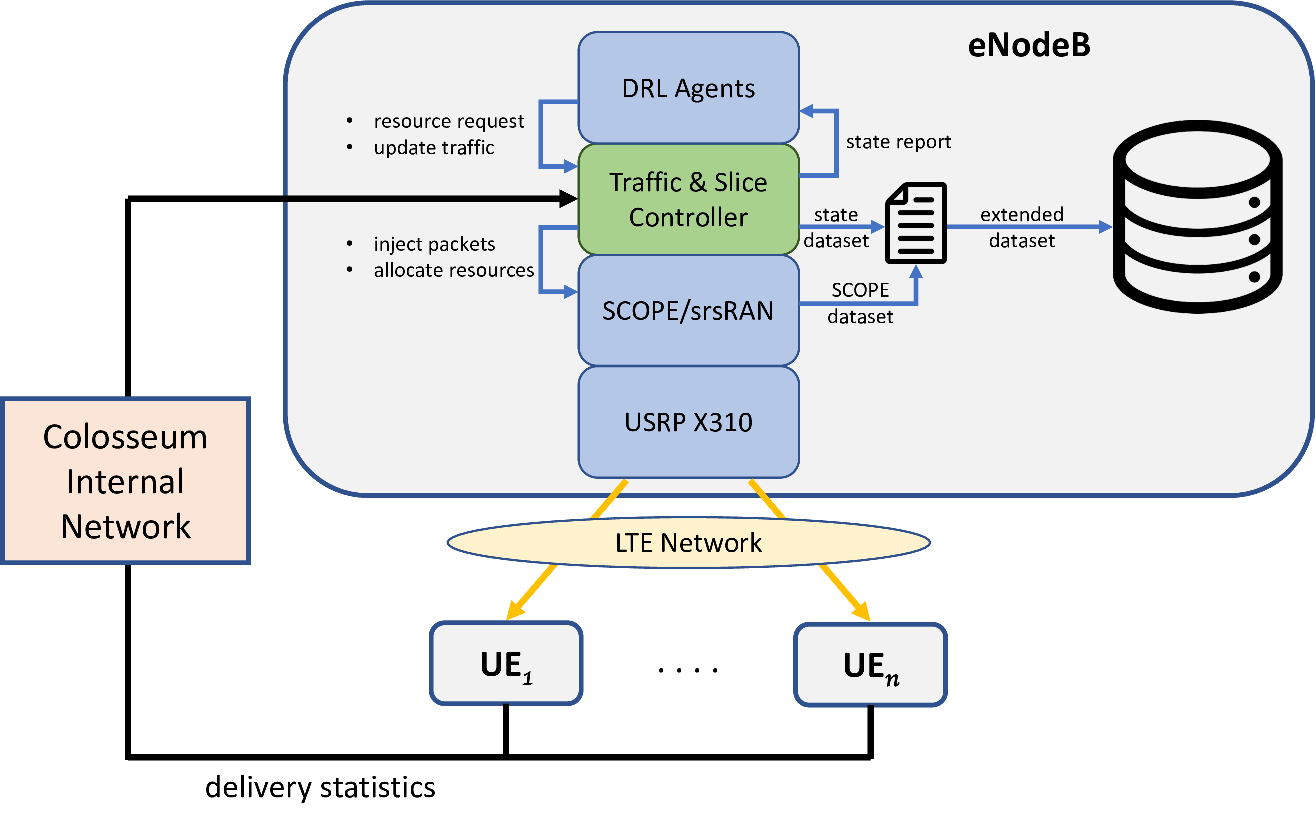}
    \caption{Experimental framework architecture.}
    \label{fig:colosseum_arch}
\end{figure}

The design of our experiment, as shown in Fig. \ref{fig:colosseum_arch}, starts with srsRAN (formerly known as srsLTE), which is an open-source implementation of the LTE and 5G standards.
The srsRAN is then run on top of the SDR devices and configured as either base stations or mobile devices, depending on the experimental scenario. To segment the srsRAN radio resources for slicing operations, we utilize SCOPE \cite{bonati2021scope}, a physical-layer slicing framework that groups the PRBs of the LTE network into separate slices. According to the LTE standard, a PRB consists of 12 subcarriers, each with a bandwidth of 15 kHz, and a duration of 0.5 ms. Hence, a single PRB has a total bandwidth of 180 kHz in frequency and a duration of 0.5 ms. By employing the SCOPE framework, we can dynamically allocate a dedicated number of PRBs to individual network slices. On top of SCOPE, we implement a custom-made slice controller that serves higher-layer applications, such as a DRL agent.

The slice controller has three primary functions: dynamically allocating the PRBs to each slice, controlling the amount of injected traffic into the LTE network based on the experimental scenario, and producing datasets at runtime for each mobile device. When invoked, the slice controller spawns eNodeBs (LTE RAN) and UEs (User Equipment). Each UE attaches itself to its assigned eNodeB via the LTE network using the SDR interface. Then each eNodeB initiates the network slices with its assigned UEs based on the initial experiment configurations. Each slice has a dedicated slice controller, which can be controlled by a dedicated DRL agent representing the MVNO discussed in previous sections of the paper.
Once a UE connects to its associated eNodeB, the UE establishes an additional connection with the eNodeB through Colosseum's internal network. This additional side channel is used to report network statistics, such as delay as shown in Fig. \ref{fig:DQN}, back to the slice controller without utilizing LTE resources used in the experiment. In other words, for each packet sent by the eNodeB, the UE sends packet statistics back to its eNodeB via the Colosseum side channel. 


These packet delivery statistics are used to generate a new system state, which is then reported back to the DRL agent every 250ms. The chosen system resolution of 250ms is limited by srsRAN, which updates its internal state four times per second. At each new system state, the DRL agent sends network traffic updates and PRB allocation requests to the slicing tool. The slicing tool updates the injection parameters of the traffic profile and reallocates the PRBs among the slices based on the agent's requests. At the end of each system state, SCOPE outputs the statistics of that state. The slice controller merges its own statistics with SCOPE's, such as the total number of PRBs used in the last 250ms. Fig. \ref{fig:colosseum_arch} provides a summary of this procedure.

\newpage

\subsection{Experimental Results} \label{sec:exp results}

Here we describe the experimental results. In the Colosseum, we used 50 available PRBs with increments/decrements of 1 resource block group (3 PRBs), setting our action set to ${-9,-6,-3,0,3,6,9}$. Due to srsRAN limitations, the agent controls the slice and computes network statistics every 250 ms. To improve sample efficiency, we employed Deep Q-learning with a learning rate of $1e-4$, a decay factor of 0.9, a batch size of 64, and an epsilon-greedy exploration strategy starting at 0.5 and decreasing to 0.05.

\begin{figure}
\centering
\begin{subfigure}{.48\linewidth}
  \centering
  \includegraphics[width=\linewidth]{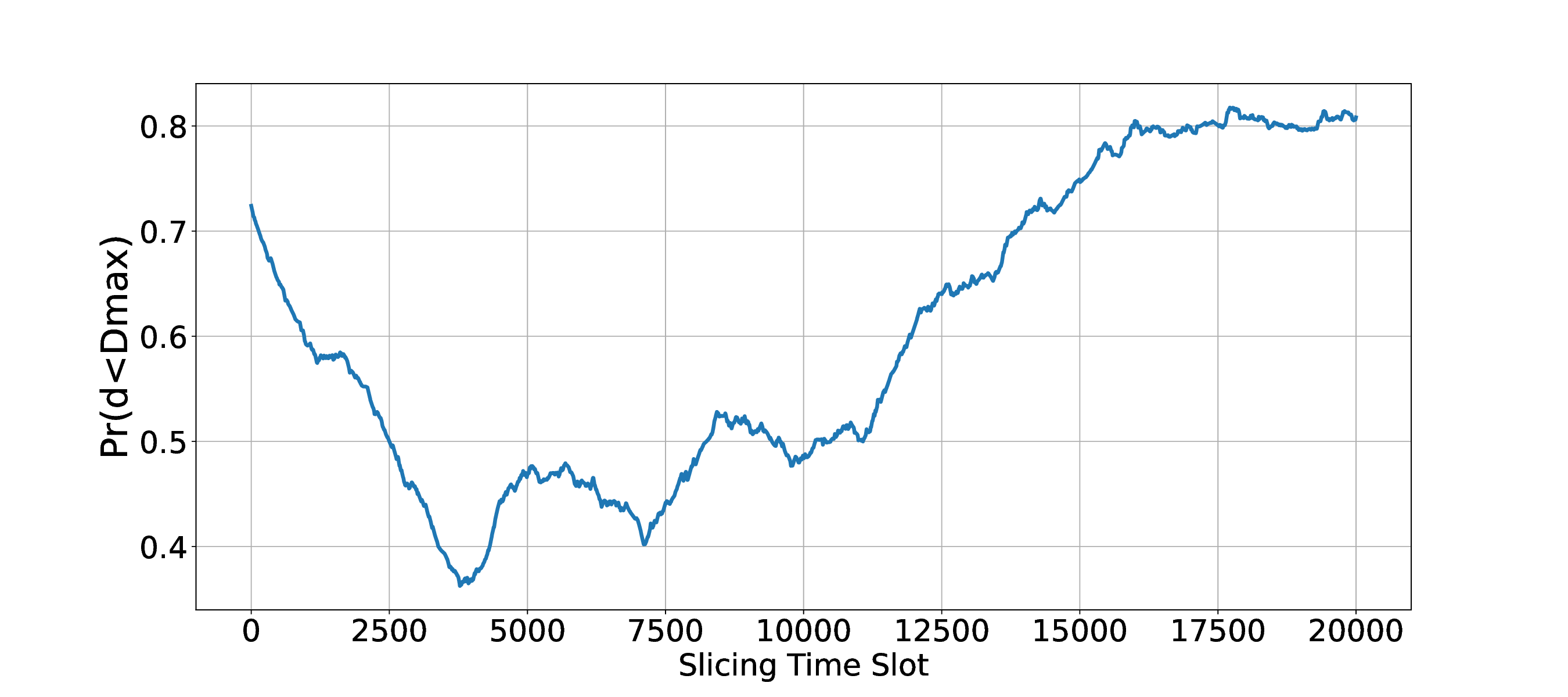}
    \caption{Probabilistic constraint.}

\end{subfigure}%
\hfill
\begin{subfigure}{.48\linewidth}
  \centering
  \includegraphics[width=\linewidth]{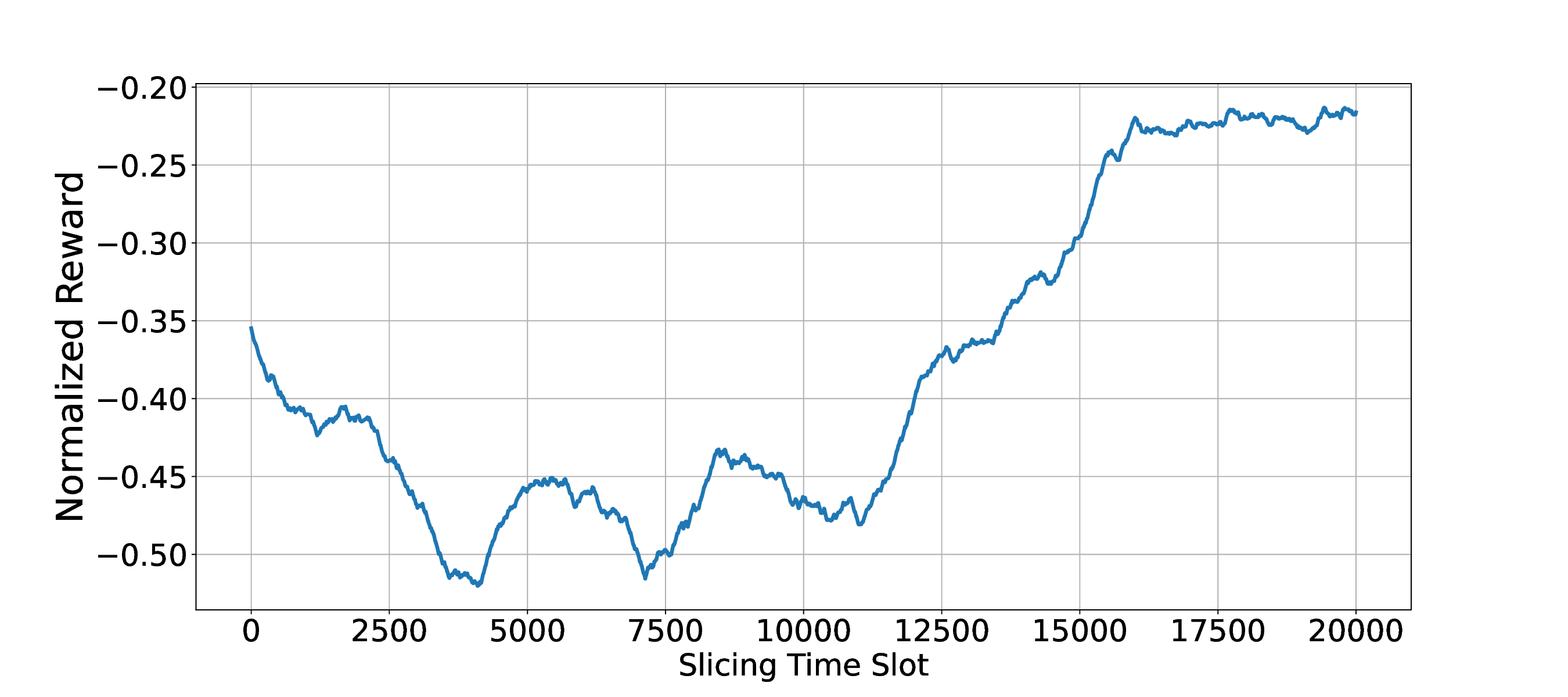}
  \caption{Normalized Reward.}
  \label{fig:sub-second}
\end{subfigure}
\vskip\baselineskip
\begin{subfigure}{.48\linewidth}
  \centering
  \includegraphics[width=\linewidth]{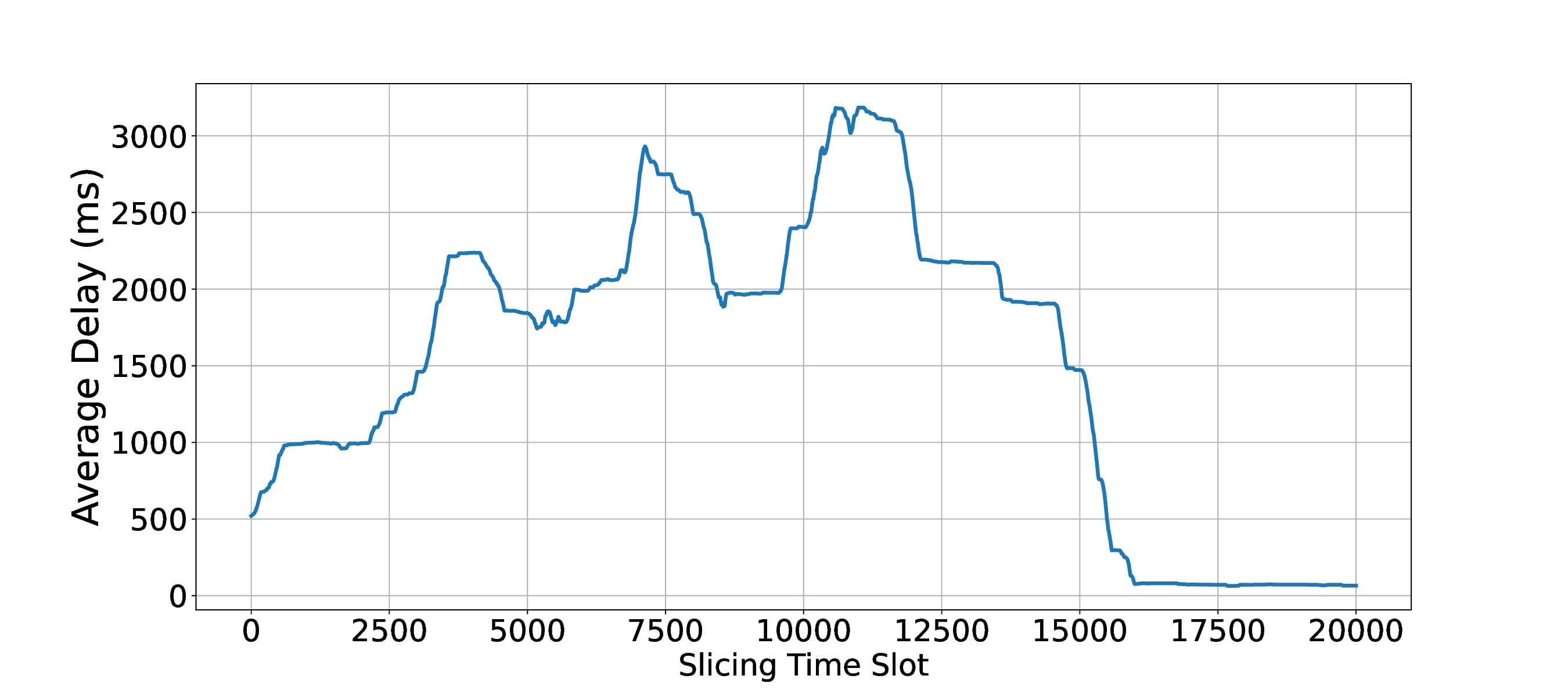}
  \caption{Average delay.}
  \label{fig:sub-third}
\end{subfigure}%
\hfill
\begin{subfigure}{.48\linewidth}
  \centering
  \includegraphics[width=\linewidth]{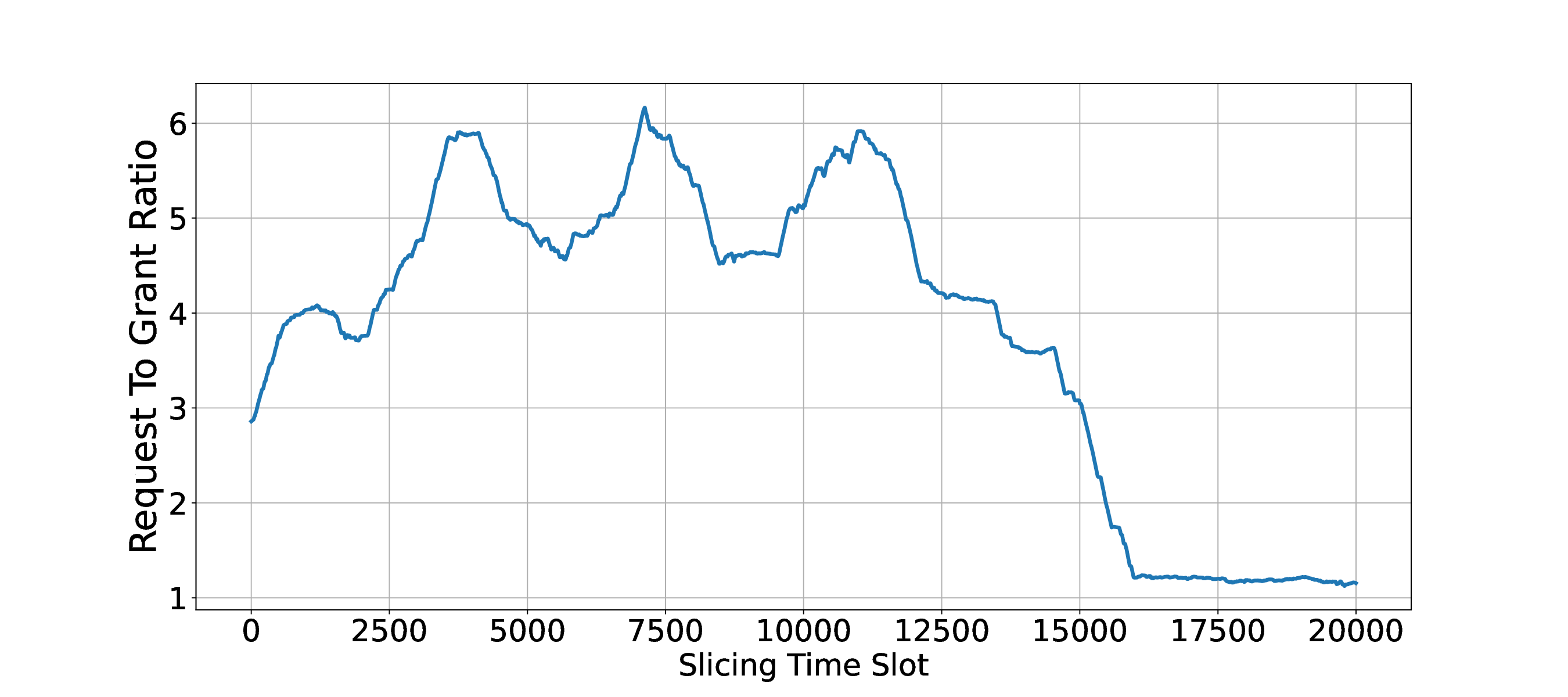}
  \caption{Requested-to-granted PRBs ratio.}
  \label{fig:sub-fourth}
\end{subfigure}
\caption{Convergence plots in experimental setting.}
\label{fig:EXP_convergence}
\end{figure}

Figure \ref{fig:EXP_convergence} shows the convergence curves during online training over 20,000 time slots, with a moving average window of 5000. The agent learns to manage PRBs to reduce delay effectively. The requested-to-granted PRBs ratio, ideally balanced around 1 to prevent excessive demand and resource wastage, converges towards this ideal by the end of training. Additionally, the average delay decreases, and the probability of satisfaction meets the constraint ($\epsilon = 0.2$).


\end{document}